\documentclass[lettersize,journal]{IEEEtran}
\IEEEoverridecommandlockouts
\usepackage{amsmath,amsfonts}
\usepackage{algorithmic}
\usepackage{array}
\usepackage[caption=false,font=normalsize,labelfont=sf,textfont=sf]{subfig}
\usepackage{textcomp}
\usepackage{stfloats}
\usepackage{url}
\usepackage{verbatim}
\usepackage{graphicx}
\def\BibTeX{{\rm B\kern-.05em{\sc i\kern-.025em b}\kern-.08em
    T\kern-.1667em\lower.7ex\hbox{E}\kern-.125emX}}
\usepackage{balance}
\usepackage{nicematrix}
\usepackage{multirow}
\usepackage{enumerate}
\usepackage[utf8]{inputenc}
\usepackage{graphicx}
\usepackage{float}
\usepackage{cite}
\usepackage{hyperref}
\usepackage{adjustbox}
\usepackage[skip=2pt]{caption} 
\usepackage{enumitem}
\usepackage{booktabs}
\usepackage{caption}
\usepackage{fancyhdr} 
\usepackage{xcolor}
\usepackage{algorithm}

\fancypagestyle{myfooter}{
  \fancyfoot[C]{\href{链接地址}{超链接内容}} 
}
\begin{document}
\title{Generative-Contrastive Heterogeneous Graph Neural Network}

\author{Yu Wang, Lei Sang*, Yi Zhang, Yiwen Zhang and Xindong Wu ~\IEEEmembership{Fellow,~IEEE}
\IEEEcompsocitemizethanks{\IEEEcompsocthanksitem Yu Wang, Lei Sang, Yi Zhang and Yiwen Zhang,  are with School of Computer Science and Technology, Anhui University 230601,
Hefei, Anhui, China.   E-mail: sanglei@ahu.edu.cn, wangyuahu@stu.ahu.edu.cn, zhangyi@stu.ahu.edu.cn, zhangyiwen@ahu.edu.cn.

\IEEEcompsocthanksitem Xindong Wu is with the Key Laboratory of Knowledge Engineering with Big Data (the Ministry of Education of China), Hefei University of Technology, Hefei 230601, Anhui, P.R. China. E-mail: xwu@hfut.edu.cn}
\thanks{*Corresponding author.}
}

\markboth{Journal of \LaTeX\ Class Files,~Vol.~14, No.~8, August~2021}%
{Shell \MakeLowercase{\textit{et al.}}: A Sample Article Using IEEEtran.cls for IEEE Journals}


\maketitle
\begin{abstract}
Heterogeneous Graphs (HGs) effectively model complex relationships in the real world through multi-type nodes and edges. In recent years, inspired by self-supervised learning (SSL), contrastive learning (CL)-based Heterogeneous Graphs Neural Networks (HGNNs) have shown great potential in utilizing data augmentation and contrastive discriminators for downstream tasks. However, data augmentation remains limited due to the graph data's integrity. Furthermore, the contrastive discriminators suffer from sampling bias and lack local heterogeneous information. To tackle the above limitations, we propose a novel \textit{Generative-Contrastive Heterogeneous Graph Neural Network (GC-HGNN)}. Specifically, we propose a heterogeneous graph generative learning method that enhances CL-based paradigm. This paradigm includes: 1) A contrastive view augmentation strategy using a masked autoencoder. 2) Position-aware and semantics-aware positive sample sampling strategy for generating hard negative samples. 3) A hierarchical contrastive learning strategy aimed at capturing local and global information. Furthermore, the hierarchical contrastive learning and sampling strategies aim to constitute an enhanced contrastive discriminator under the generative-contrastive perspective. Finally, we compare our model with seventeen baselines on eight real-world datasets. Our model outperforms the latest baselines on node classification and link prediction tasks. To reproduce our work, we have open-sourced our code at https://github.com/wangyu0627/GC-HGNN.
\end{abstract}


\section{Introduction}
\IEEEPARstart{H}{eterogeneous} Information Networks (HIN) or Heterogeneous Graphs (HG) \cite{2013hin,2019han} have emerged as critical tools for analyzing the real world, as they capture rich semantic information by modeling different types of nodes and edges. This is particularly relevant in networks such as citation networks \cite{2017metapath2vec, 2019han}, recommender systems \cite{2018herec,2023hgcl}, and social networks \cite{2021smin}. Recently, Heterogeneous Graph Neural Networks (HGNNs) \cite{2019han,2021heco} demonstrate a strong ability in processing HIN data because they expand the receptive field of nodes through meta-path \cite{2011meta-path} and message-passing mechanism, thereby capturing the rich semantics.

\begin{figure}[t]
    \centering
    \includegraphics[width=0.95\linewidth]{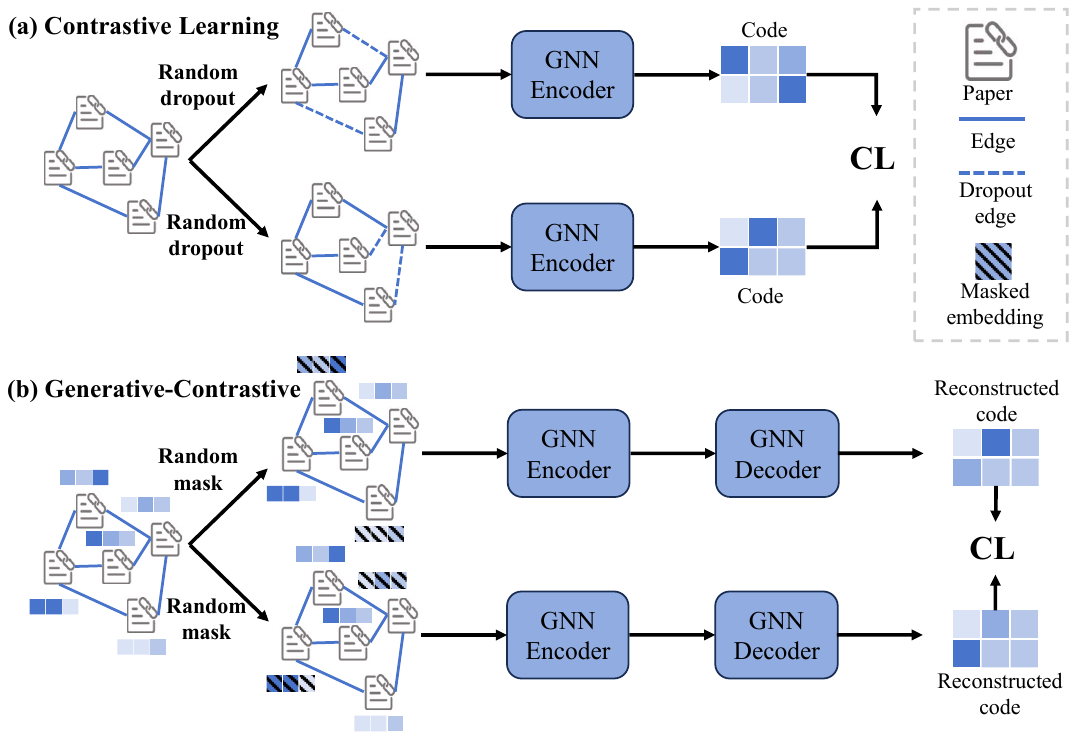}
    \caption{(a) Contrastive Learning paradigm aligns two similar views constructed through random edge dropping; (b) Generative-Contrastive paradigm aligns two reconstructed embeddings without altering the original graph structure.}
    \label{fig0:motivation}
\end{figure}

Self-supervised learning (SSL) has been widely applied in the fields of Computer Vision (CV) \cite{2020moco,2020byol} and Natural Language Processing (NLP) \cite{2023palm,2018bert} leveraging unlabeled data to learn informative representations. Inspired by this trend, researchers explore the combination of HGNNs and SSL. As a major paradigm, HGNNs \cite{2021heco, 2022sehgcl} based on Contrastive Learning (CL) show great potential, primarily due to their suitability for downstream tasks achieved by setting contrastive views and discriminators, as demonstrated in \cite{2020moco, 2021ssl}. Data augmentation \cite{2021sgl, 2023vgcl} serves as a tool to construct construct contrastive views, maximizing mutual information by aligning two views. However, given the requirement to maintain graph data's integrity and unique features, it is challenging to provide reasonable data augmentation to leverage the benefits of contrastive learning.

Generative learning \cite{2023palm, 2018bert} effectively addresses this limitation by reconstructing the data distribution, disregarding downstream tasks and data augmentation, thereby enhancing each node embedding. This idea further serves as a foundation for proposing a generative learning enhanced contrastive learning paradigm for heterogeneous graphs. As shown in Figure \ref{fig0:motivation}, first, HGNNs utilize a meta-path to obtain a graph containing nodes of the same type. CL-based HGNN \cite{2023hgcml, 2022sehgcl} generates supervised signals by aligning two similar views constructed through random edge dropping; meanwhile, generative-contrastive learning \cite{2020gan, 2023gacn} reconstructs the intrinsic structures and features of data by training an autoencoder and uses a contrastive discriminator to distinguish them in the embedding space. Despite the obvious advantages of the generative-contrastive paradigm, it presents several challenges for heterogeneous graphs:

\noindent \textbf{CH1: Unbalanced data augmentation.}
The current graph augmentation strategies \cite{2023hgcml, 2021sgl}, such as structural and feature augmentation, still have limitations. On the one hand, structural augmentation \cite{2023hgcml, 2022sehgcl}, which involves randomly dropping out nodes and edges, might disrupt the graph integrity, as nodes and edges are the fundamental properties of a graph. On the other hand, feature augmentation \cite{2021heco, 2022ncl}, which perturbs node attributes, tends to overlook the unique features of each node's input. In these methods, certain nodes have more edges than others, which results in unbalanced supervised signals. Exploring how to design enhanced contrastive views represents a challenge that requires attention.

\noindent \textbf{CH2: Negative sampling bias.}
In heterogeneous graphs, contrastive learning \cite{2021heco} heavily relies on high-quality negative samples. Moreover, the contrastive discriminator requires stricter sampling methods under the generative-contrastive paradigm \cite{2023gacn}. Existing sampling strategies \cite{2023hgcml, 2021heco} are typically based on randomness or meta-paths, and these approaches frequently lead to an increase in false negative samples. Recent works \cite{2020hard,2023hard} aim to select or synthesize hard negatives to mitigate the impact of false negatives. However, they are not suitable for HGs and provide limited benefits. Therefore, there is an urgent need to design an enhanced sampling strategy for HGs.

\noindent \textbf{CH3: Lack local heterogeneous information.}
Existing methods \cite{2023hgcml, 2023hgmae} rely solely on meta-paths to capture rich semantic information in HGs and often overlook the influence of one-hop neighbors on target nodes. Network schema leverages direct connections between nodes to capture the local structure of target nodes. In contrast, meta-paths are employed to extract higher-order structures. Designing an effective self-supervised mechanism that considers both local and global information can significantly enhance the model's performance.

To address the three challenges mentioned above, we propose a heterogeneous graph self-supervised model called Generative-Contrastive Heterogeneous Graph Neural Network (GC-HGNN). 
For \textbf{CH1}, GC-HGNN employs a generative masked autoencoder (MAE) to enhance contrastive views. The MAE is capable of reconstructing node embeddings without altering the original graph structure and features. The masking strategy averages the edges of each node, ensuring that each node receives the same supervised signals during the reconstruction process. Moreover, random structural perturbations in existing methods can exacerbate imbalance, potentially leading to adverse effects on node representations \cite{2023vgcl}. Our proposed GC-HGNN uses an autoencoder to create embeddings without modifying the graph's topology or inherent node features, preserving useful information. The proposed sampling strategy is not simply "node drop" but a selective uniform random sampling without replacement. This method helps prevent bias by ensuring neighbors are neither fully masked nor entirely visible, which enhances feature restoration \cite{gilmer2017neural}.
For \textbf{CH2}, the model introduces a sampling strategy that integrates both location-awareness and semantic-awareness. This approach adaptively samples nodes based on the rich semantics inherent to HGs and the global positions of nodes within these graphs. In contrastive learning, the goal is to minimize positive sample distances while maximizing negative ones. In GC-HGNN, the positive sampling strategy also improves hard negative selection, promoting feature similarity in positive pairs while reducing similarity in negative pairs. Studies show that suboptimal false negatives can lead to erroneous guidance \cite{wei2024llmrec}, and precise positive sampling enhances discrimination, thus indirectly supporting negative sample quality \cite{zhuo2024improving}.
Finally, for \textbf{CH3}, GC-HGNN employs hierarchical contrastive learning (HCL) to capture both one-hop and higher-order neighbors. Intra-contrast is employed to contrast different meta-path views, while inter-contrast is used to contrast between network schema and meta-path views. Our model builds upon HeCo’s framework but simplifies and generalizes it by encoding all edge types directly, which minimizes preprocessing cost and avoids semantic loss. Unlike HeCo’s two-step attention aggregation, our approach is more streamlined.

We summarize the contributions of this paper as follows:

\begin{itemize}[leftmargin=*]
\item We propose a generative learning method that enhances contrastive learning (CL)-based paradigm GC-HGNN, leveraging a generative masked autoencoder for view augmentation. This represents a significant exploration for graph generative-contrastive networks.
\item We introduce hierarchical contrastive learning (HCL) as an enhanced contrastive discriminator, which is integrated with innovative sampling strategy to align the generative model. These modules effectively mine self-supervised signals.
\item We conduct comprehensive experiments to evaluate GC-HGNN with seventeen baseline methods and eight real datasets. The experimental results demonstrate that GC-HGNN is superior to baselines and effectively extracts heterogeneous information.
\end{itemize}

\section*{Related Work}
This section examines the development process of heterogeneous graph neural networks. Here, we introduce the issue of infeasible and expensive data labels, further expanding on the self-supervised advantages and providing a brief overview of existing models.

\subsection{Heterogeneous Graph Neural Networks}
In recent years, Graph Neural Networks (GNNs) \cite{2020graphGCL, 2022graphmae, 2017gat, 2020lightgcn} have demonstrated significant success in fields such as social network analysis and recommender systems. This success is attributed to their effective modeling of graph structure information and the integration of both local and global information. These graph structures typically have different types of nodes and edges. Heterogeneous graph neural networks (HGNNs) are designed to better handle heterogeneous information.

The currently prevalent HGNN-based models \cite{sang2024denoising, 2021heco, 2019hetgnn, 2020magnn, 2022rhgnn} can be categorized into meta-path-based methods and type-based methods. The classic meta-path-based models include HAN \cite{2019han}, HetGNN \cite{2019hetgnn}, and MAGNN \cite{2020magnn}. Specifically, HAN employs a dual-layer attention mechanism at both the node and semantic levels to adaptively capture the relationships between nodes and meta-paths. HetGNN introduces a heterogeneous graph attention propagation mechanism, combining meta-path information and attention mechanisms to learn node representations. MAGNN integrates meta-paths and adaptive mechanisms to learn node representations in heterogeneous graphs by learning meta-path attention weights at both the node and global levels. The type-based models include RGCN \cite{2018rgcn}, HGT \cite{2020hgt}, and RHGNN \cite{2022rhgnn}. RGCN captures rich semantic relationships by using different weight matrices for different types of edges. HGT achieves efficient representation learning and information propagation on heterogeneous graph data by introducing a multi-head attention mechanism and a heterogeneous graph structure encoder. RHGNN utilizes dedicated graph convolution components and cross-relation passing modules to learn node representations on heterogeneous graphs. However, the methods mentioned above focus exclusively on supervised heterogeneous graph models. \textcolor{black}{FrameERC \cite{review3_1, review3_3} leverages the Framelet Transform in a multimodal GNN for emotion recognition in conversations.} In order to deeply explore the potential relationships and patterns in the data and to effectively utilize unlabeled data, an increasing number of studies are beginning to focus on self-supervised learning methods. 

\subsection{Self-supervised Graph Learning}
Given the rapid development of self-supervised learning in computer vision \cite{2020moco,2020simclr} and natural language processing \cite{2018bert,2023palm} fields, graph neural networks have benefited from this progress. Self-supervised graph neural networks typically train without labels by generating contrastive views or reconstructing node attributes, thereby better leveraging the rich structural information in graph data. Transitioning from homogeneous graphs to heterogeneous graphs entails assigning different types or attributes to nodes and edges in the graph, rendering the graph structure more complex and diverse in representation.

Self-supervised heterogeneous graph neural networks can be categorized into contrastive and generative. Contrastive learning models \cite{2019hdgi,2021heco,2023hgcml} generate enhanced contrastive views and aim to maximize the mutual information between these views to produce supervised signals. Specifically, HDGI \cite{2019hdgi} utilizes graph convolutional modules and semantic-level attention to learn node representations and maximize local-global mutual information. HeCo \cite{2021heco} complements across perspectives through network schema and meta-path views, establishing a sampling contrast method between the two views to achieve self-supervised enhancement. HGCML \cite{2023hgcml} solely utilizes the rich semantic relationships brought by meta-paths for graph enhancement, introducing multi-view contrastive learning. It also proposes sampling strategies for positive samples in both topology and semantics to alleviate sampling bias. Generative models \cite{sang2024intent,2023hgmae} commonly adopt graph autoencoder and aim to reconstruct features to obtain supervision. For example, MvDGAE \cite{2021MvDGAE} utilizes random edge dropout to enhance views and autoencoders to enforce the recovery of views for more robust representations. Meanwhile, HGMAE \cite{2023hgmae} further introduces a dynamic masking mechanism and position-aware encoding to enhance the model's performance and utilizes attribute and view dual-recovery to improve denoising capability.

\begin{figure}[t]
    \centering
    \includegraphics[width=0.9\linewidth]{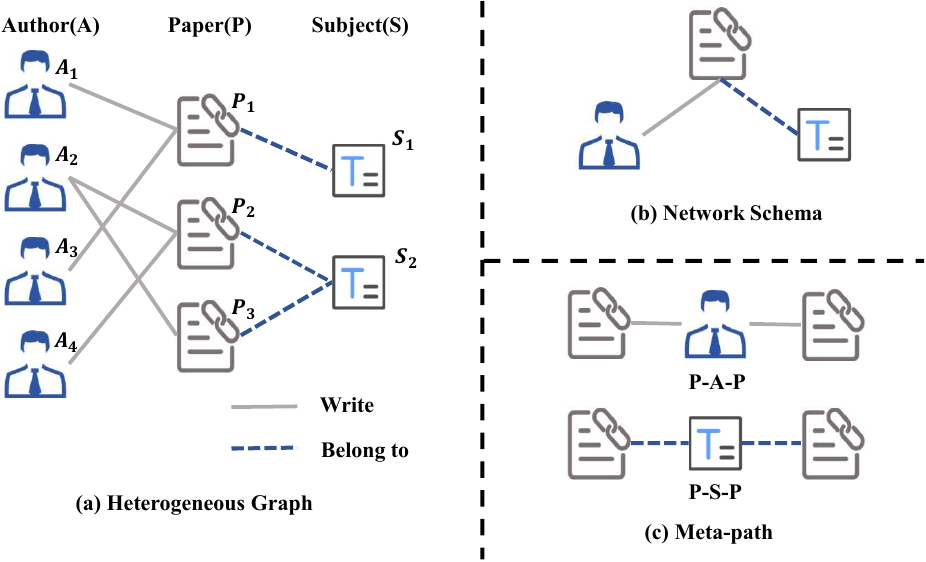}
    \caption{A toy example of an HIN on ACM dataset.}
    \label{fig1:intro}
\end{figure}

\begin{table}[h!]
    \centering
    \caption{\textcolor{black}{Summary of symbols and notations.}}
    \color{black} 
    \begin{tabular}{cl}
        \toprule
        \textbf{Symbol} & \textbf{Description} \\
        \midrule
        $\mathcal{G}$ & A heterogeneous information network (HIN) \\
        $V$, $E$ & Sets of nodes and edges \\
        $I$, $J$ & Sets of node types and edge types \\
        $T_G$ & Network schema of the HIN \\
        $\rho_j$ & A meta-path in the set of selected meta-paths \\
        $\boldsymbol{P}_{r}$ & Projection matrix \\
        $\alpha_{e w}$ & Attention weight between node $e$ and neighbor $w$ \\
        $\sigma$ & Activation function \\
        $\boldsymbol{h}_{e}^{1}$ & Updated embedding of node $e$ in network schema \\
        $G=\mathcal{V}, \mathcal{A}, \boldsymbol{X}$ & Definition of a meta-path-based-subgraph with \\
                                                     & nodes, adjacency matrix, and feature matrix \\
        $\mathcal{V}^{[Mask]}$ & Set of masked nodes \\
        $p$ & Mask ratio for masked nodes \\
        $\widetilde{\boldsymbol{X}}$ & Masked node feature matrix \\
        $\mathcal{V}^{[Remask]}$ & Set of remasked nodes \\
        $\boldsymbol{H}, \widetilde{\boldsymbol{H}}, \boldsymbol{H}_{Re}$ & Encoded feature, decoder input with the remasked \\
                                                                          & feature matrix, reconstructed output \\
        $w_{\rho_{j}}$ & Weight score for meta-path $\rho_{j}$ \\
        $\gamma_{\rho_{j}}$ & Attention weight for meta-path $\rho_{j}$ \\
        $\boldsymbol{h}_{e}^{2}$ & Updated embedding of node $e$ in meta-path view \\
        $\mathcal{E}$ & Edge sampling set for inter-contrast \\
        $\widetilde{\mathcal{V}}$ & Positive sample set for intra-contrast \\
        $\mathbb{P}_{i}$, $\mathbb{N}_{i}$ & Positive and negative sample sets for inter-contrast \\
        $\tau$ & Temperature parameter in contrastive loss \\
        $\lambda$ & Balance coefficient of contrastive loss $\mathcal{L}_{Inter}$ \\
        $\mathcal{L}_{Gen/Intra/Inter}$ & Loss functions of generative learning, \\
                                        & intra-contrast and inter-contrast \\
        $\mathcal{L}_{bpr}$ & Bayesian personalized ranking loss \\
        \bottomrule
    \end{tabular}
    \label{table:notation}
\end{table}

\section{PRELIMINARIES}
In this section, we formally define some significant concepts related to HIN as follows:

\noindent \textbf{Heterogeneous Information Network (HIN):} A HIN is characterized by a graph structure $\mathcal{G}=(V, E, I, J, \phi, \varphi)$, in which the collections of nodes and edges are symbolized by $V$ and $E$ respectively. For every node $v$ and edge $e$, there are associated type mapping functions $\phi: V \rightarrow I$ for nodes and $\varphi: E \rightarrow J$ for edges. Here, $I$ represents the node types and $J$ denotes the edge types, with the total number of types exceeding two, i.e., $|I|+|J|>2$.

\noindent \textbf{Network Schema:} The network schema, denoted as $T_{G}=(I, J)$, is used to demonstrate the high-level topology of an HIN, which describes the node types and their interaction relations. The network schema itself is a directed graph that is predicated on a collection of node types $I$, where the links between them are represented by the set of relational edge types $J$. The network schema for the ACM is depicted in the top portion of Figure \ref{fig1:intro} (b).

\noindent \textbf{Meta-path:} In an HIN $\mathcal{G}=(V, E, I, J, \phi, \varphi)$, a meta-path $\rho$ is depicted as $I_{1} \stackrel{J_{1}}{\longrightarrow} I_{2} \stackrel{J_{2}}{\longrightarrow} \ldots \stackrel{J_{l}}{\longrightarrow} I_{l+1}$, which describes a composite connection between $I_{1}$ and $I_{l+1}$. Taking Figure \ref{fig1:intro} as an instance, multiple meta-paths, such as ``Paper-Author-Paper (PAP)'' and ``Paper-Subject-Paper (PSP)'' can link entities within the HIN. Typically, different meta-paths can reveal the varying inter-dependency information between two papers. The meta-path ``PAP'' indicates that two papers were written by the same author, while ``PSP'' illustrates that two papers belong to the same subject.
\textcolor{black}{Before introducing the proposed model, we summarize the symbols used in Table \ref{table:notation} in this paper.}

\begin{figure*}[t]
    \centering
    \includegraphics[width=0.9\linewidth]{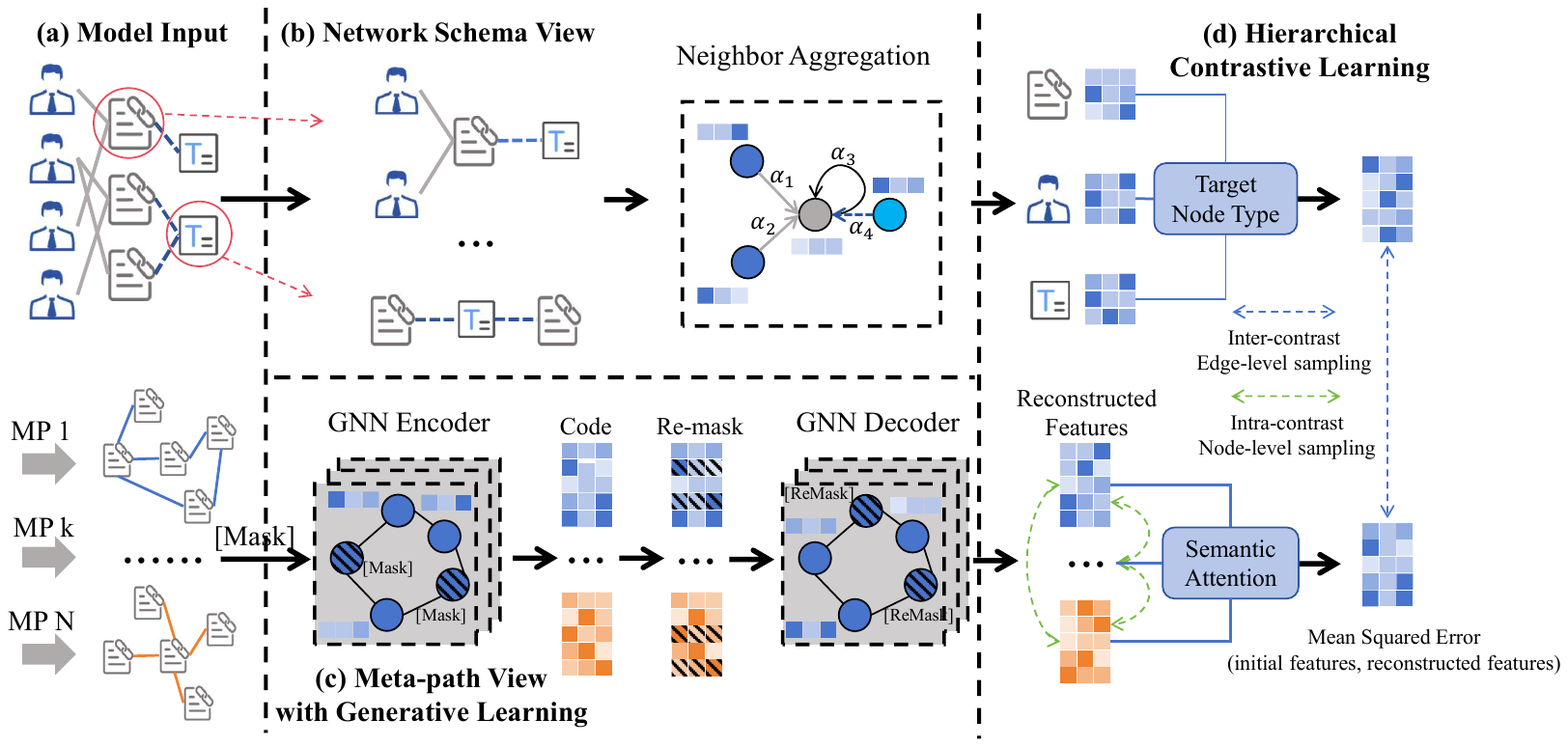}
    \caption{The training framework of our proposed GC-HGNN. (a) The model respectively takes the entire heterogeneous graph and the views based on meta-paths, dividing them into inputs for the network schema View and the meta-path view; (b) the network schema view employs an attention mechanism to focus on one-hop neighbors of each node; (c) the meta-path view utilizes a masked autoencoder to generate views for intra-contrast; (d) the hierarchical contrast uses a proposed dynamic sampling mechanism to generate more hard negative samples to enhance the contrastive discriminator.}
    \label{fig2:model}
\end{figure*}

\section{THE PROPOSED MODEL}
In this section, we present overall framework and brief module descriptions, as shown in Figure \ref{fig2:model}. We then provide detailed explanations for each module following the framework's order. Finally, we propose different optimization strategies based on the downstream tasks.
\subsection{Network Schema View}
This section takes the entire heterogeneous information network as input, as shown in Figure \ref{fig2:model} (a). First, we consider each node in the network as an anchor, forming a subnetwork that encompasses all the one-hop neighbor nodes. For example, in Figure \ref{fig2:model} (b), the subgraph indicated by the circles represents the subnetwork formed by the target node. Next, we input such subnetworks into a heterogeneous neighbor aggregation network, utilizing attention mechanisms to obtain node embeddings. Finally, we select the embeddings of the desired target node for downstream tasks. \textbf{Specific implementation process:} We project all nodes into the same node space ($\boldsymbol{P}_{r}$ is projection matrix) due to the various node types within the subnetworks. We define $e$ as the current node computing the attention score, and $\boldsymbol{h}_{e}^{0}$ as its initial embedding. Next, we learn aggregation functions for different types of edges to compute attention scores:
\begin{equation}
\alpha_{e w}=\frac{\exp \left(f_{r}\left(\boldsymbol{h}_{e}^{0}, \boldsymbol{P}_{r} \boldsymbol{h}_{w}^{0}\right)\right)}{\sum_{j \in \mathcal{N}_{1}(e)} \exp \left(f_{r}\left(\boldsymbol{h}_{e}^{0}, \boldsymbol{P}_{r} \boldsymbol{h}_{j}^{0}\right)\right)}, \quad \forall w \in \mathcal{N}_{1}(e) \\
\label{eq1}
\end{equation}
where $f_{r}(\cdot, \cdot)$ denotes the deep neural network designed to implement attention mechanisms over one-hop neighbors. Here, $\alpha_{ew}$ is the influence level for neighbor node $w$, and $\mathcal{N}_{1}(e)$ denotes the set of one-hop neighbor nodes of $e$. Finally, we aggregate the following information from $\mathcal{N}_{1}(e)$: 
\begin{equation}
\boldsymbol{h}_{e}^{1}=\sigma\left(\sum_{w \in \mathcal{N}_{1}(e)} \alpha_{e w} \boldsymbol{h}_{w}^{0}\right)
\label{eq2}
\end{equation}
where $\sigma$ denotes the activation function.

\subsection{Meta-path View with Generative Learning} \label{4.2}
In this section, the model inputs are multi-views constructed from different meta-paths. To address the issue of imbalanced data augmentation, we enhance multiple views using a generative mask autoencoder. For example, in Figure \ref{fig2:model} (a), we connect two nodes of the same type via the specified meta-path `PAP'. Thus, such a graph composed of nodes of the same type is termed a view based on the meta-path `PAP'. Based on the preliminaries, we define a set of meta-paths as $\left\{\rho_{1}, \ldots, \rho_{j}, \ldots, \rho_{M}\right\}$, and the corresponding views as $\left\{G^{\rho_1}, G^{\rho_2}, \ldots, G^{\rho_M}\right\}$. $M$ denotes the number of meta-paths we have selected. Here, the model defines a graph $G=(\mathcal{V}, \mathcal{A}, \boldsymbol{X})$. For each graph, $\mathcal{V}$ is the node set, $N=$ $|\mathcal{V}|$ is the number of nodes, $\mathcal{A} \in\{0,1\}^{N \times N}$ is the adjacency matrix, and $\mathbf{X} \in \mathbb{R}^{N \times d}$ is the input node feature matrix.

First, we employ a sampling strategy without replacement to select a set of nodes $\mathcal{V}^{[Mask]}$ from the entire nodes set $\mathcal{V}$: 
\begin{equation}
    \mathcal{V}^{[Mask]} = \{ \mathcal{V}_{i} \in \mathcal{V} \mid r_i \leq p \}, 
    r_i \sim \text{Uniform}(0, 1)
\end{equation}
where $p$ is the mask ratio we set, and $\mathcal{V}_{i}$ is the node set based on $r_i$. The uniform random sampling \cite{2022graphmae} helps the encoder avoid potential bias centers, which refers to the bias arising when all neighbors of a node are either completely masked or completely visible. We mask the entire node features in the set $\mathcal{V}^{[Mask]}$ to obtain $\widetilde{\boldsymbol{X}}$.

Then, we input the masked features into a GNN encoder to obtain the output. To better reconstruct the features, we set the encoder with adjustable layers $L$. A multi-layer encoder typically captures distant neighbors to expand their receptive field, enriching their representations.
\begin{equation}
\boldsymbol{H}=f_{E}^{(L)}(\mathcal{A}, \widetilde{\boldsymbol{X}})
\end{equation}
where $\widetilde{\boldsymbol{X}}$ is feature input with the $[Mask]$, and $\boldsymbol{H} \in \mathbb{R}^{N \times d}$ denotes code after the encoder. Next, to further encourage the decoder to learn compressed representations, we use a new ratio and the same sampling strategy to mask a different set of nodes $\mathcal{V}^{[ReMask]}$. We use a GNN with the same structure as the encoder to act as the decoder. However, the decoder can be GAT \cite{2017gat}, GCN \cite{2020lightgcn}, and its layer number and type can be different.
\begin{equation}
\boldsymbol{H}_{Re}=f_{D}^{(L)}(\mathcal{A}, \widetilde{\boldsymbol{H}})
\end{equation}
where $\widetilde{\boldsymbol{H}}$ is decoder input with the $[Remask]$, and $\boldsymbol{H}_{Re}$ as the reconstructed output under meta-path view. We perform multiple masking operations between encoders, equivalent to the masked nodes reconstructing their features through their neighbors. The uniform sampling strategy ensures that each reconstructed node receives the same supervised signals. Such an autoencoder helps train robust node embeddings and mitigates the impact of noise.

Finally, based on the mentioned meta-path-based multi-view, we denote the reconstructed features obtained from the different views as $\left\{\boldsymbol{H}_{Re}^{\rho_1}, \ldots, \boldsymbol{H}_{Re}^{\rho_j}, \ldots, \boldsymbol{H}_{Re}^{\rho_M}\right\}$. We employ a semantic attention mechanism to assign weights to each view to balance their importance in downstream tasks.
\begin{equation}
w_{\rho_{j}}=\frac{1}{|\mathcal{V}|} \sum_{\boldsymbol{e} \in \mathcal{V}} \boldsymbol{q}^{T} \cdot \tanh \left(\boldsymbol{W} {\boldsymbol{h}_{Re}^{\rho_{j}}}+b\right),
\end{equation}
where ${\boldsymbol{h}_{Re}^{\rho_{j}}}$ is a reconstructed feature of a node under meta-path $\rho_{j}$. $\boldsymbol{q}^{T}$ and $\boldsymbol{W}$ are query vectors and weight matrices to ensure that the dimensions and feature spaces match different views, respectively. Next, the weight for $\rho_{j},(j = 1: N)$ is defined as:
\begin{equation}
\gamma_{\rho_{j}}=\frac{\exp \left(w_{\rho_{j}}\right)}{\sum_{j=1}^{N} \exp \left(w_{\rho_{j}}\right)},
\end{equation}
where $\gamma_{\rho_{j}}$ denotes the weight of meta-path $\rho_{j}$. According to the attention mechanism \cite{2019han}, when the semantic attention score $\gamma_{\rho_{j}}$ is higher, it indicates the meta-path $\rho_{j}$ gets more importance in downstream tasks. We aggregate feature embeddings from all meta-paths to obtain the final node features under the meta-path view.
\begin{equation}
\boldsymbol{h}_{e}^{2}=\sum_{j=1}^{N} \gamma_{\rho_{j}} \cdot \boldsymbol{h}_{e}^{\rho_{j}}
\end{equation}

The key idea of generative learning lies in feature reconstruction. Therefore, we fit the reconstructed masked node features $\boldsymbol{h}_{e}^{2}$ and initial features $\boldsymbol{h}_{e}^{0}$, using the Mean Squared Error (MSE) \cite{2020gptgnn} as the loss function:
\begin{equation}
\mathcal{L}_{Gen} = \frac{1}{|\mathcal{V}^{[Mask]}|} \sum_{e \in \mathcal{V}^{[Mask]}} \left(\boldsymbol{h}_{e}^{0} - \boldsymbol{h}_{e}^{2}\right)^2
\end{equation}
where $e$ denotes the node in the masked set. Previous GAEs \cite{2016vgae} typically used a single-layer MLP or Laplacian matrix in decoding and focused more on restoring the graph structure.

\begin{figure}[t]
    \centering
    \includegraphics[width=0.95\linewidth]{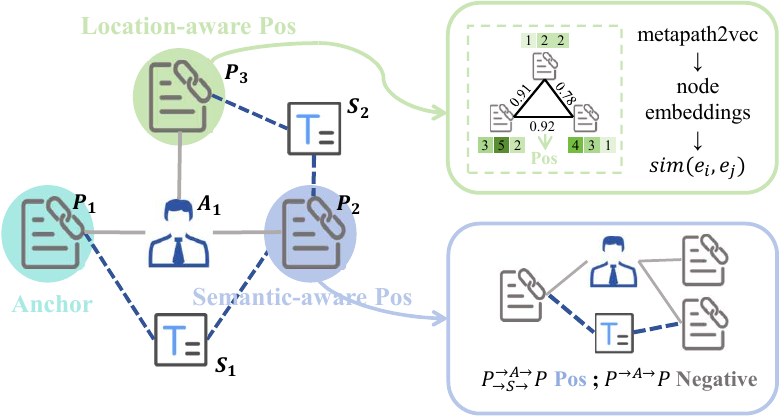}
    \caption{Our GC-HGNN proposed a hard sampling strategy for inter-contrast. Location-aware Pos: the green node uses metapath2vec \cite{2017metapath2vec} to perceive the location within a heterogeneous graph to obtain node embeddings. Semantic-aware Pos: The blue node is connected to the anchor node through all types of meta-paths, indicating the semantic similarity between them.}
    \label{fig3:sample}
\end{figure}

\subsection{Hierarchical Contrastive Learning}
The inputs for this section are the embeddings derived from the network schema and meta-path, denoted as $\boldsymbol{h}{e}^{1}$ and $\boldsymbol{h}{e}^{2}$. To mitigate sampling bias and information loss within the contrastive discriminator, we introduce innovative sampling strategies and a hierarchical contrast framework. The model utilizes two distinct sampling strategies: node-level contrast for intra-contrast and edge-level contrast for inter-contrast. The process of edge-level sampling is depicted in Figure \ref{fig3:sample}. Subsequently, we obtain an edge sampling set, denoted as $\mathcal{E}$.

\textcolor{black}{To obtain high-quality positive samples, we propose two sampling strategies: \textbf{Location-aware Pos} and \textbf{Semantic-aware Pos}, which automatically extract positive samples from the data. As shown in Figure \ref{fig3:sample}, the green nodes represent Location-aware Pos, while the blue nodes denote Semantic-aware Pos.} 
\begin{itemize}[leftmargin=*]
    \item \textcolor{black}{Location-aware Pos calculates the similarity between ``$P_1-P_3$'' using node embeddings pre-trained by Metapath2vec \cite{2017metapath2vec} and selects the top-k nodes to complete the sampling process. Metapath2vec employs meta-path-based random walks to effectively capture structural information in HGs.}
    \item \textcolor{black}{Semantic-aware Pos connects ``$P_1-P_2$'' through two meta-paths, ``$P_1-S_1-P_2$'' (PSP) and ``$P_1-A_1-P_2$'' (PAP), rather than a single meta-path. The more meta-path connections exist, the higher the semantic similarity between them.}
\end{itemize}
\textcolor{black}{We chose metapath2vec for position-aware positives as it captures structural information through meta-path-based walks, making it better suited for long-sequence link awareness. Relying solely on model-based embeddings would weaken position awareness and lose distant node relations \cite{2023hgmae}.}

\noindent \textbf{Intra-contrast.}
First, we utilize sampling set $\mathcal{E}$ to select the positive sample set $\widetilde{\mathcal{V}}$ for node-level contrast based on the degree of nodes. High-degree nodes have more connections in the graph and are more likely to capture the central structure and important patterns in the graph. The embeddings of the model, based on multi-views, are denoted by $\left\{\boldsymbol{H}_{Re}^{\rho_1}, \ldots, \boldsymbol{H}_{Re}^{\rho_j}, \ldots, \boldsymbol{H}_{Re}^{\rho_M}\right\}$. Next, we take these embeddings as input to obtain a multi-view loss. Intra-contrast employs Scaled Cosine Error (SCE) \cite{2020byol} as the contrastive loss between multi-views. SCE has been proven to focus on hard samples in the data to improve the stability of representation learning. 

\begin{equation}
\mathcal{L}_{Intra}=\frac{1}{|\widetilde{\mathcal{V}}|} \sum_{v_{i} \in \widetilde{\mathcal{V}}}\left(1-\frac{\boldsymbol{x}_{i}^{T} \boldsymbol{z}_{i}}{\left\|\boldsymbol{x}_{i}\right\| \cdot\left\|\boldsymbol{z}_{i}\right\|}\right)^{\eta}, \eta \geq 1
\label{eq10}
\end{equation}
where $\boldsymbol{x}$ and $\boldsymbol{z}$ respectively denote embeddings of the same node under two meta-path views. During training, we can scale the cosine error by a power of $\eta \geq 1$ to minimize the differences of important samples under multi-views.

\noindent \textbf{Inter-contrast.}
Inter-contrast performs collaborative contrastive learning between two non-aligned views. In set $\mathcal{E}$, we define the target node to be sampled as $i$ and the connected nodes to it as a positive set $\mathbb{P}_{i}$, inspired by \cite{2021heco}. Otherwise, they form the negative sample set $\mathbb{N}_{i}$. From the scales of $\mathcal{E}$ and $\widetilde{\mathcal{V}}$, inter-contrast typically has more hard negative samples, and thus, its supervisory role is more evident. The contrastive loss of the network schema view with respect to the meta-path view is as follows:
\begin{equation}
\mathcal{L}_{Inter}^{1}=-\log \frac{\sum_{j \in \mathbb{P}_{i}} \exp \left(\operatorname{sim}\left(h_{i}^{1}, h_{j}^{2}\right) / \tau\right)}{\sum_{k \in\left\{\mathbb{P}_{i} \cup \mathbb{N}_{i}\right\}} \exp \left(\operatorname{sim}\left(h_{i}^{1}, h_{k}^{2}\right) / \tau\right)}
\label{eq11}
\end{equation}
where $\operatorname{sim}(i,j)$ is the cosine similarity, and $\tau$ denotes a temperature parameter. 

Inter-contrast utilizes mutual contrast between the two views to achieve the collaborative process. The contrastive loss of the meta-path view with respect to the network schema view is $\mathcal{L}_{Inter}^{2}$. The overall inter-contrast losses are as follows:
\begin{equation}
\mathcal{L}_{{Inter}}=\frac{1}{|\mathcal{E}|} \sum_{i \in \mathcal{E}}\left[\lambda \cdot \mathcal{L}_{Inter}^{1}+(1-\lambda) \cdot \mathcal{L}_{Inter}^{2}\right]
\label{eq12}
\end{equation}
where $\lambda$ is a balance coefficient to control the weight of each view pair. We employ intra-contrast and inter-contrast for hierarchical contrastive learning and analyze their operational mechanisms. The intra-contrast assumes that meta-paths are interconnected and aims to maximize their mutual information \cite{2021ssl}. This method integrates the different semantics across meta-paths to form a generalized knowledge representation for downstream tasks. On the other hand, the inter-contrast approach attempts to enhance node representation by contrasting local and global information, preventing nodes from being influenced by noise from higher-order neighbors. The hierarchical contrast and sampling strategies are designed to construct a contrastive discriminator to enhance the adversarial interplay between the generative and contrastive models.

\subsection{Model Optimization}\label{4.4}
In this subsection, we describe the optimization strategies for different tasks in our experiments. It has two aspects: node classification and link prediction.

\noindent \textbf{Node classification.}
First, the common practice for the node classification task \cite{2020dmgi, 2021heco} is to train embeddings for all nodes using self-supervised loss. Then, we fine-tune the model on different datasets using a few labeled samples and a single-layer MLP. The node-level and edge-level strategies we propose typically lead to considerable loss differences. Therefore, it is necessary to balance their weights during the training phase. The loss generated by hierarchical contrastive learning is as follows:
\begin{equation}
\mathcal{L}_{hcl} = \lambda_{Intra} \cdot \mathcal{L}_{{Intra}} + \lambda_{Inter} \cdot \mathcal{L}_{{Inter}} + \lambda_{Gen} \cdot \mathcal{L}_{{Gen}}
\end{equation}
where $\lambda_{{Intra}}$, $\lambda_{{Inter}}$ and $\lambda_{Gen}$ are hyper-parameters to balance the influences of losses.

\noindent \textbf{Link prediction.}
Next, for the link prediction task \cite{2023hgcl,2018herec}, we introduce the Bayesian Personalized Ranking Loss (BPR) \cite{2012bpr}. This loss function treats two connected nodes as a positive pair and two unconnected nodes as a negative pair. Its objective is to maximize the scores between positive pairs and negative pairs. Formally, the BPR loss is defined as follows:
\begin{equation}
\mathcal{L}_{b p r}=-\frac{1}{|\mathcal{O}|} \sum_{(i, j, k) \in O} \log \sigma\left(\mathbf{d}_{i}^{\top} \mathbf{d}_{j}-\mathbf{d}_{i}^{\top} \mathbf{d}_{k}\right)
\end{equation}
where $\mathcal{O}=\left\{(i, j, k) \mid A_{i, j}=1, A_{i, k}=0\right\}$ is the training data. We combine $\mathcal{L}_{{hcl}}$ and $\mathcal{L}_{bpr}$ to get the loss:
\begin{equation}
\mathcal{L} = \lambda_{hcl} \cdot \mathcal{L}_{{hcl}} + \lambda_{bpr} \cdot \mathcal{L}_{{bpr}}
\end{equation}
where $\lambda_{hcl}$ and $\lambda_{bpr}$ denote hyperparameters used to balance task weights.

\subsection{Time Complexity Analysis}
The network schema view's time complexity is $O\left(|V| \cdot D_n d^2\right)$, where $|V|$ denotes the edge number of a heterogeneous graph. $D_n$ is the average node degree and $d$ is the feature dimension. For the meta-path view, GC-HGNN has $M$ meta-path-based subgraphs. So, the time complexity of the meta-path view is $O\left(|\mathcal{V}| \cdot 2 M L D d\right)$, where $L$ is the number of GNN layers. Then, the self-supervised loss contains $\lambda_{Intra}$ and $\lambda_{Inter}$. The time complexity to get positive samples and calculate $\lambda_{Intra}$ is $O\left(|\widetilde{\mathcal{V}}| \cdot d\right)$. For the $\lambda_{Inter}$, the complexity is $O\left(|\mathcal{V}|^2 \cdot d\right)$. Because the edge set $\mathcal{E}$ is a matrix $\mathcal{E} \in\{0,1\}^{N \times N}$ composed of target nodes. In the link prediction task, the complexity of the additional $\mathcal{L}_{{bpr}}$ is $O\left(|\mathcal{O}| \cdot d\right)$. Thus, the overall time complexity of GC-HGNN is $O\left(\left(|V| \cdot D_n d + |\mathcal{V}| \cdot 2 M L D_m + |\widetilde{\mathcal{V}}| + |\mathcal{V}|^2 + |\mathcal{O}|\right) \cdot d \right)$. This complexity is the time complexity for each epoch during training, and the number of training epochs will determine the overall training time.
We compared several strong baselines in terms of training time. Our model performs faster per epoch due to optimized modules, demonstrating lower complexity and greater scalability compared to baseline methods.

\begin{table}[t]
    \centering
    \captionsetup{justification=centering}
    \caption{Time and resource comparison on datasets.}
    \begin{adjustbox}{width=0.4\textwidth}
    \begin{NiceTabular}{l c c c c}
        \midrule
        \textbf{Dataset} & \textbf{ACM} & \textbf{DBLP} & \textbf{Aminer} & \textbf{Freebase} \\
        \midrule
        \multicolumn{5}{c}{\textbf{Time} (eopch/s)} \\ \midrule
        HeCo \cite{2021heco}    & 0.204 & 0.198 & 0.361 & 0.272 \\
        HGMAE \cite{2023hgmae}   & 0.212 & 0.334 & 0.203 & 0.114 \\
        GC-HGNN  & \textbf{0.033} & \textbf{0.058} & \textbf{0.043} & \textbf{0.037} \\
        \midrule
        \multicolumn{5}{c}{\textbf{Resource} (memory/G)} \\ \midrule
        HeCo \cite{2021heco} & \textbf{2.21} & \textbf{3.25} & 8.75 & \textbf{6.36} \\
        HGMAE \cite{2023hgmae} & 5.49 & 6.82 & 14.48 & 8.93 \\
        GC-HGNN  & 2.54 & 4.10 & \textbf{8.33} & 6.65 \\
        \midrule
    \end{NiceTabular}
    \end{adjustbox}
    \label{table0:complexity}
\end{table}

\begin{table}[t]
    \captionsetup{justification=centering}
    \caption{Statistics of datasets used in this paper.}
    \begin{adjustbox}{width=0.48\textwidth}
    \begin{NiceTabular}{lccccc}
    \midrule Dataset & Nodes & Edges & Relations & Task \\
    \midrule ACM & 11,246 & 34,852 & 2 & Node Classification \\
    DBLP & 26,128 & 239,566 & 3 & Node Classification \\
    Aminer & 55,783 & 153,676 & 2 & Node Classification \\
    Freebase & 43,854 & 151,043 & 3 & Node Classification \\
    \midrule Last.fm & 31,469 & 913,216 & 3 & Link Prediction \\
    Yelp & 31,092 & 976,276 & 5 & Link Prediction \\
    Douban Movie & 37,595 & 3,429,882 & 6 & Link Prediction \\
    Douban Book & 51,029 & 4,451,894 & 7 & Link Prediction \\
    \midrule
    \end{NiceTabular}
    \end{adjustbox}
    \label{table1:datasets}
\end{table}

\section{EXPERIMENT}
In this section, we conduct extensive experiments and answer the following research questions:
\begin{itemize}[leftmargin=*]
\item \textbf{RQ1:} How does GC-HGNN perform w.r.t. tasks of node classification and link prediction?
\item \textbf{RQ2:} Are the key components in our GC-HGNN delivering the expected performance gains?
\item \textbf{RQ3:} How does GC-HGNN combine contrast and generation to achieve performance beyond one or the other?
\item \textbf{RQ4:} How do different settings influence the effectiveness of GC-HGNN?
\end{itemize}

\begin{table*}[t]
    \centering
    \captionsetup{justification=centering}
    \caption{Quantitative results (\%) on node classification. In the table, the symbols $\mathbf{X}$, $\mathbf{A}$, and $\mathbf{Y}$ denote the node features, graph structure, and predicted node labels, respectively.}
    \begin{adjustbox}{width=\textwidth}
    \begin{NiceTabular}{c|c|cccccccc}
    \toprule[1pt]
    \midrule \multirow{2}{*}{ Methods } & \multirow{2}{*}{ Data } & \multicolumn{2}{c}{ ACM } & \multicolumn{2}{c}{ DBLP } & \multicolumn{2}{c}{ Aminer } & \multicolumn{2}{c}{ Freebase } \\
    \cmidrule{3-10} & & Macro-F1 & Micro-F1 & Macro-F1 & Micro-F1 & Macro-F1 & Micro-F1 & Macro-F1 & Micro-F1 \\
    \midrule DeepWalk(2017) & $\mathbf{A} $ & $68.34 \pm 0.42$ & $74.05 \pm 0.47$ & $89.24 \pm 0.49$ & $90.12 \pm 0.45$ & $68.48 \pm 0.38$ & $74.00 \pm 0.41$ & $59.20 \pm 0.50$ & $59.85 \pm 0.43$ \\
    MP2vec(2017) & $\mathbf{A}$ & $68.05 \pm 0.98$ & $72.86 \pm 0.92$ & $89.98 \pm 0.84$ & $90.30 \pm 0.38$ & $66.28 \pm 1.17$ & $72.21 \pm 0.84$ & $55.50 \pm 0.49$ & $53.17 \pm 0.34$ \\
    HERec(2018) & $\mathbf{A}$ & $68.84 \pm 0.44$ & $74.45 \pm 0.45$ & $89.44 \pm 0.67$ & $90.37 \pm 0.47$ & $68.73 \pm 0.83$ & $74.21 \pm 0.87$ & $59.46 \pm 1.16$ & $60.11 \pm 0.46$ \\
    \midrule HAN(2019) & $\mathbf{X}, \mathbf{A}, \mathbf{Y}$ & $88.30 \pm 0.23$ & $88.20 \pm 0.37$ & $86.00 \pm 0.53$ & $85.18 \pm 0.49$ & $64.15 \pm 0.35$ & $75.60 \pm 0.79$ & $58.50 \pm 0.93$ & $61.30\pm 0.41$ \\
    HGT(2020) & $\mathbf{X}, \mathbf{A}, \mathbf{Y}$ & $87.09 \pm 0.35$ & $88.20 \pm 0.26$ & $91.36 \pm 0.23$ & $91.24 \pm 0.49$ & $64.15 \pm 0.35$ & $75.60 \pm 0.24$ & $58.30 \pm 0.72$ & $61.34 \pm 0.18$ \\
    MHGCN(2022) & $\mathbf{X}, \mathbf{A}, \mathbf{Y}$ & $88.86 \pm 0.63$ & $88.70 \pm 0.22$ & $90.62 \pm 0.58$ & $90.62 \pm 0.19$ & $71.34 \pm 0.10$ & $79.24 \pm 0.79$ & $57.12 \pm 0.32$ & $59.30 \pm 0.33$ \\
    \midrule DMGI(2020)& $\mathbf{X}, \mathbf{A}$ & $87.66 \pm 0.24$ & $87.60 \pm 0.35$ & $89.55 \pm 0.52$ & $90.66 \pm 0.46$ & $63.15 \pm 0.21$ & $66.61 \pm 0.38$ & $55.99 \pm 0.59$ & $59.37 \pm 0.27$ \\
    HeCo(2021) & $\mathbf{X}, \mathbf{A}$ & $88.56 \pm 0.28$ & $88.71 \pm 0.43$ & $90.26 \pm 0.37$ & $90.59 \pm 0.25$ & $71.64 \pm 0.22$ & \underline{$79.95 \pm 0.55$} & $59.87 \pm 0.33$ & $62.33 \pm 0.40$ \\
    HGCML(2023) & $\mathbf{X}, \mathbf{A}$ & $88.46 \pm 0.38$ & $87.90 \pm 0.27$ & $90.47 \pm 0.33$ & $91.11 \pm 0.49$ & $71.84 \pm 0.45$ & $78.30 \pm 0.49$ & $58.92 \pm 2.57$ & $62.71 \pm 0.23$ \\
    HGMAE(2023) & $\mathbf{X}, \mathbf{A}$ & \underline{$89.29 \pm 0.36$} & \underline{$89.15 \pm 0.31$} & \underline{$91.60 \pm 0.58$} & \underline{$91.89 \pm 0.35$} & \underline{$72.63 \pm 0.27$} & $79.87 \pm 0.42$ & \underline{$60.82 \pm 0.51$} & \underline{$63.89 \pm 0.38$} \\
    \midrule GC-HGNN & $\mathbf{X}, \mathbf{A}$ & $\mathbf{90.54} \pm \mathbf{0.28}$ & $\mathbf{90.36} \pm \mathbf{0.31}$ & $\mathbf{92.54} \pm \mathbf{0.36}$ & $\mathbf{93.01} \pm \mathbf{0.30}$ & $\mathbf{73.44} \pm \mathbf{0.20}$ & $\mathbf{81.28} \pm \mathbf{0.32}$ & $\mathbf{62.47} \pm \mathbf{0.42}$ & $\mathbf{65.85} \pm \mathbf{0.35}$ \\
    \midrule \bottomrule
    \end{NiceTabular}
    \end{adjustbox}
    \label{table2:node results}
\end{table*}

\subsection{Experimental Setings}
We followed the experimental benchmarks of HeCo and HGMAE, where the hidden layer dimension is fixed at 128, and the number of fine-tuning samples per class is set to 20, 40, and 60. The original HGMAE paper used a hidden layer dimension of 512, for which we readjusted the optimal parameters. Due to space limitations, we reported the results for 20 samples, as labels are usually valuable, and fewer samples often require stronger model generalization ability. All experimental results were repeated 10 times on the 4090, and we release the reproducible framework for the model as open source at the code\footnote{\url{https://github.com/wangyu0627/HGNN-baselines}}.

\subsection{Datasets}
The performance of GC-HGNN has been validated on six real-world datasets. These datasets encompass four datasets for node classification: ACM, DBLP, Aminer, and Freebase, as well as four datasets for link prediction: Last.fm, Yelp, Douban Movie and Douban Book. A statistical overview of all the datasets is presented in Table \ref{table1:datasets}.

\noindent \textbf{Datasets for Node Classification.}
\begin{itemize}[leftmargin=*]
\item \textbf{ACM} \cite{2022openhgnn}. The target nodes are papers (3 classes), averagely linked to 3.33 authors and one subject.
\item \textbf{DBLP} \cite{2020magnn}. The target nodes are authors (4 classes), averagely linked to 4.84 papers.
\item \textbf{Freebase} \cite{2023hgmae}. The target nodes are movies (3 classes), which are typically linked to an average of 18.7 actors, 1.07 directors, and 1.83 writers.
\item \textbf{Aminer} \cite{2021heco}. The target nodes are papers (4 classes), averagely linked to 2.74 authors and 8.96 references.
\end{itemize}
\noindent \textbf{Datasets for Link Prediction.}
\begin{itemize}[leftmargin=*]
\item \textbf{Last.fm}\footnote{\url{https://www.last.fm/}} predicts user-artist interactions in a music recommendation setting, with other node type including tag.
\item \textbf{Yelp}\footnote{\url{https://www.yelp.com/}} predicts user-business interactions in a business recommendation setting, with node types including category, city and category,etc.
\item \textbf{Douban Movie}\footnote{\url{https://m.douban.com/}} predicts user-movie interactions in a movie recommendation setting, with node types including actor, director and type, etc.
\item \textbf{Douban Book} \cite{2018herec} predicts user-book interactions in a book recommendation setting, with node types including author, publisher and group, etc.
\end{itemize}
\subsection{Baselines}
We respectively select the most representative baseline model for node classification and link prediction.

\noindent \textbf{Heterogeneous Graph Representation Learning Models.}
\begin{itemize}[leftmargin=*]
\item \textbf{DeepWalk} \cite{2014deepwalk} maps network structure to vector space via random walks, preserving neighbor similarity.
\item \textbf{MP2vec} \cite{2017metapath2vec} employs meta-path-guided random walks for heterogeneous network embeddings.
\item\textbf{HERec} \cite{2018herec} integrates meta-path-based random walks for node embeddings in HINs with fusion functions.
\item \textbf{HAN} \cite{2019han} employs node-level and semantic-level attention to capture node and meta-path importance.
\item \textbf{HGT} \cite{2020hgt} introduces a novel architecture for handling heterogeneous structures in Web-scale graphs.
\item \textbf{DMGI} \cite{2020dmgi} achieves self-supervised learning by meta-path-based networks and mutual infromation between networks.
\item \textbf{HeCo} \cite{2021heco} uses collaborative supervised contrast between network schema view and meta-path view.
\item \textbf{MHGCN} \cite{2022mhgcn} utilizes multi-layer convolution aggregation to capture meta-path interactions of different lengths.
\item \textbf{HGCML} \cite{2023hgcml} uses meta-paths to generate multi-views combined with comparative learning between views.
\item \textbf{HGMAE} \cite{2023hgmae} captures graph information by reconstructing masked meta-path-based view and embedding.
\end{itemize}
\noindent \textbf{Graph Recommendation Models.}
\begin{itemize}[leftmargin=*]
\item \textbf{BPRMF} \cite{2012bpr} learns implicit feature vectors between users and items to maximize difference in pairwise preferences.
\item \textbf{LightGCN} \cite{2020lightgcn} removes the feature transformation and nonlinear activations of GCN to achieve recommendation.
\item \textbf{SGL} \cite{2021sgl} generates contrast views by edge dropout to aid contrastive learning to enhance recommendation.
\item \textbf{SMIN} \cite{2021smin} utilizes metagraph informax network to enhance user preference representation for social recommendation.
\item \textbf{NCL} \cite{2022ncl} considers the relationship between neighbor nodes to enhance collaborative filtering.
\item \textbf{HGCL} \cite{2023hgcl} integrates heterogeneous relational semantics by leveraging contrastive self-supervised learning.
\item \textbf{VGCL} \cite{2023vgcl} introduces a variational graph reconstruction for graph augmentation to achieve robust recommendation.
\end{itemize}

\begin{table*}[t]
    \centering
    \captionsetup{justification=centering}
    \caption{Quantitative results on link prediction.}
    \begin{adjustbox}{width=0.9\textwidth}
    \begin{NiceTabular}{c|cccccccc}
    \toprule[1pt]
    \midrule Dataset & \multicolumn{2}{c}{ Douban-Book } & \multicolumn{2}{c}{ Douban-Movie } & \multicolumn{2}{c}{ Last.fm } & \multicolumn{2}{c}{ Yelp } \\
    \midrule Metric & Recall@20 & NDCG@20 & Recall@20 & NDCG@20 & Recall@20 & NDCG@20 & Recall@20 & NDCG@20 \\
    \midrule BPRMF(2012) & 0.1302 & 0.1112 & 0.1678 & 0.1888 & 0.2400 & 0.2349 & 0.0752 & 0.0466 \\
    HERec(2018) & 0.1313 & 0.1201 & 0.1714 & 0.1919 & 0.2526 & 0.2484 & 0.0822 & 0.0471 \\
    LightGCN (2020) & 0.1412 & 0.1307 & 0.1791 & 0.2011 & 0.2623 & 0.2638 & 0.0893 & 0.0565 \\
    SGL-ED (2021) & 0.1685 & \underline{0.1631} & 0.1890 & 0.2056 & 0.2724 & 0.2704 & 0.0971 & 0.0614 \\
    NCL (2022) & 0.1694 & 0.1623 & 0.1910 & 0.2074 & 0.2738 & 0.2718 & 0.0975 & \underline{0.0623}  \\
    VGCL (2023) & \underline{0.1701} & 0.1602 & \underline{0.1933} & \underline{0.2097} & \underline{0.2740} & \underline{0.2726}& \underline{0.0978} & 0.0621 \\
    \midrule HAN (2019) & 0.1290 & 0.1137 & 0.1676 & 0.1884 & 0.2443 & 0.2397 & 0.0767 & 0.0449 \\
    HGT (2020) & 0.1308 & 0.1157 & 0.1696 & 0.1904 & 0.2463 & 0.2417 & 0.0787 & 0.0469 \\
    SMIN (2021) & 0.1353 & 0.1239 & 0.1718 & 0.1909 & 0.2547 & 0.2560 & 0.0847 & 0.0485 \\
    HGCL (2023) & 0.1526 & 0.1426 & 0.1868 & 0.2024 & 0.2684 & 0.2688 & 0.0954 & 0.0607 \\
    \midrule GC-HGNN & $\mathbf{0.1890}$ & $\mathbf{0.1717}$ & $\mathbf{0.1997}$ & $\mathbf{0.2175}$ & $\mathbf{0.2813}$ & $\mathbf{0.2795}$ & $\mathbf{0.1044}$ & $\mathbf{0.0649}$ \\
    Improv. & $11.11 \%$ & $7.18 \%$ & $3.31 \%$ & $3.72 \%$ & $2.63 \%$ & $2.53 \%$ & $6.75 \%$ & $4.51 \%$ \\
    \midrule \bottomrule[1pt]
    \end{NiceTabular}
    \end{adjustbox}
    \label{table3:link results}
\end{table*}

\subsection{Overall Performance (RQ1)}
\noindent \textbf{Node Classification.}
The experiment results are shown in Table \ref{table2:node results}. Notably, self-supervised methods generally outperform unsupervised ones based on shallow networks. This performance disparity arises because unsupervised models only consider the graph structure, whereas self-supervised models endeavor to mine hidden semantic relationships from the HIN. Conversely, the performance of supervised models is dependent on more labels, which goes against the challenge of label scarcity and preciousness. HGMAE, a generative reconstruction model, demonstrates the significant potential of autoencoders in graph representation learning, while HeCo, a contrastive model, also yields remarkable results. These two models learn graph representations from different perspectives, providing insights for designing future GNN models. Inspired by these insights, GC-HGNN integrates the generative masked autoencoder with a contrastive discriminator to improve performance by generating more hard negative samples.

\noindent \textbf{Link Prediction.}
Recommendation is a classic task for heterogeneous link prediction. Considering the task adaptability, we use LightGCN \cite{2020lightgcn} to replace the network schema. Here, we standardize the embedding dimension to 64 and select the recall and NDCG \cite{zhang2023revisiting} as evaluation metrics. The results are shown in Table \ref{table3:link results}. It can be observed that NCL and HGCL, as representatives of contrastive learning, demonstrate distinct advantages over other models within their respective categories. The essence of contrastive learning lies in achieving uniformity and alignment. However, modifying the graph structure to achieve uniformity is time-consuming and tricky, so it is better to consider this aspect from the perspective of the embedding space. Our GC-HGNN incorporates embeddings learned from the meta-path view into the network schema view and performs hierarchical contrastive learning. Finally, across four datasets, our model achieved improvements of over 3\% compared to the strongest baseline models.

\begin{figure}[t]
    \centering
    \includegraphics[width=\linewidth]{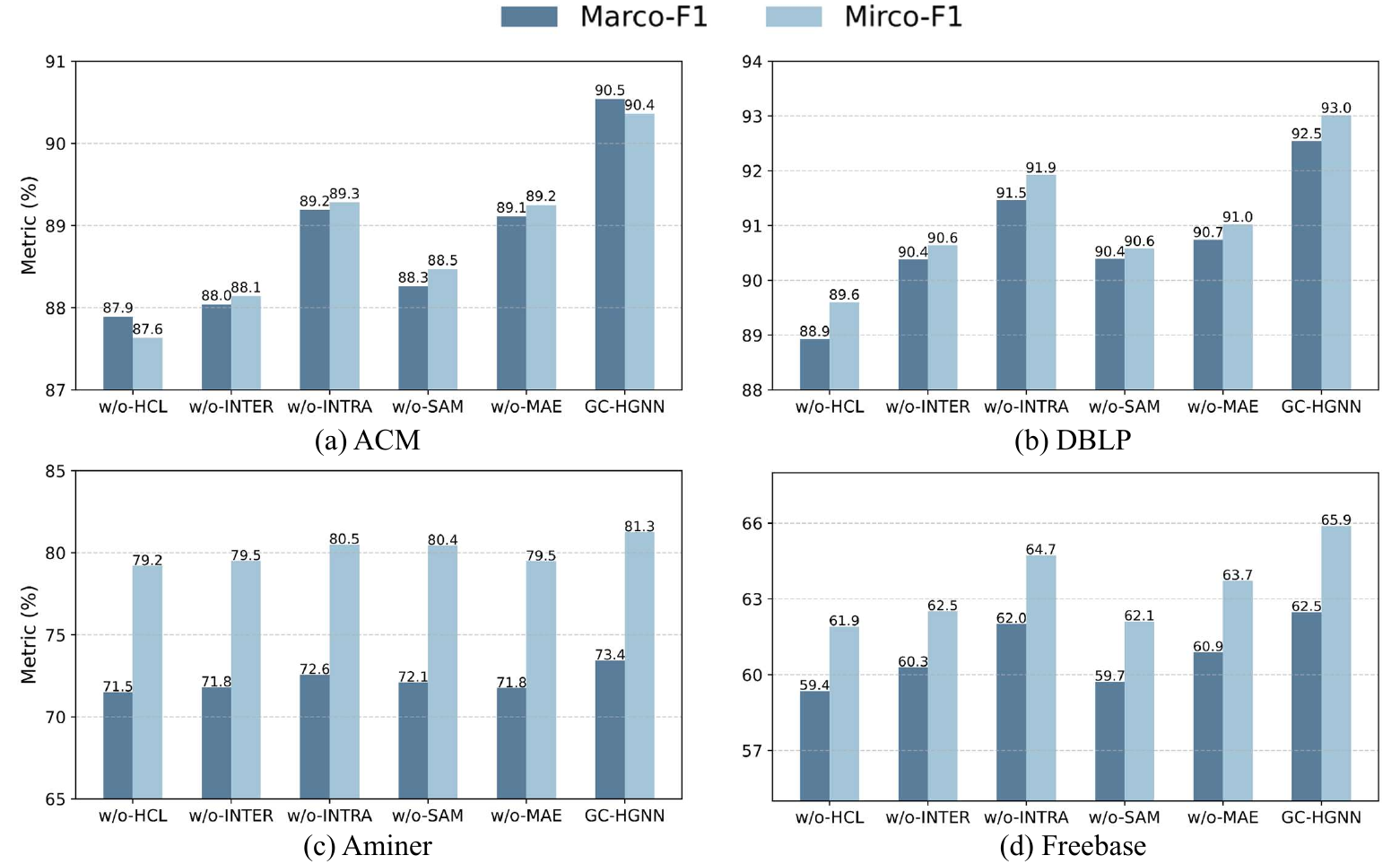}
    \caption{\textcolor{black}{The comparison of GC-HGNN and its variants.}}
    \label{fig4:ablation}
\end{figure}

\begin{figure}[t]
    \centering
    \begin{minipage}[b]{0.45\linewidth}
        \centering
        \subfloat[\small MP2vec]{\includegraphics[width=\linewidth]{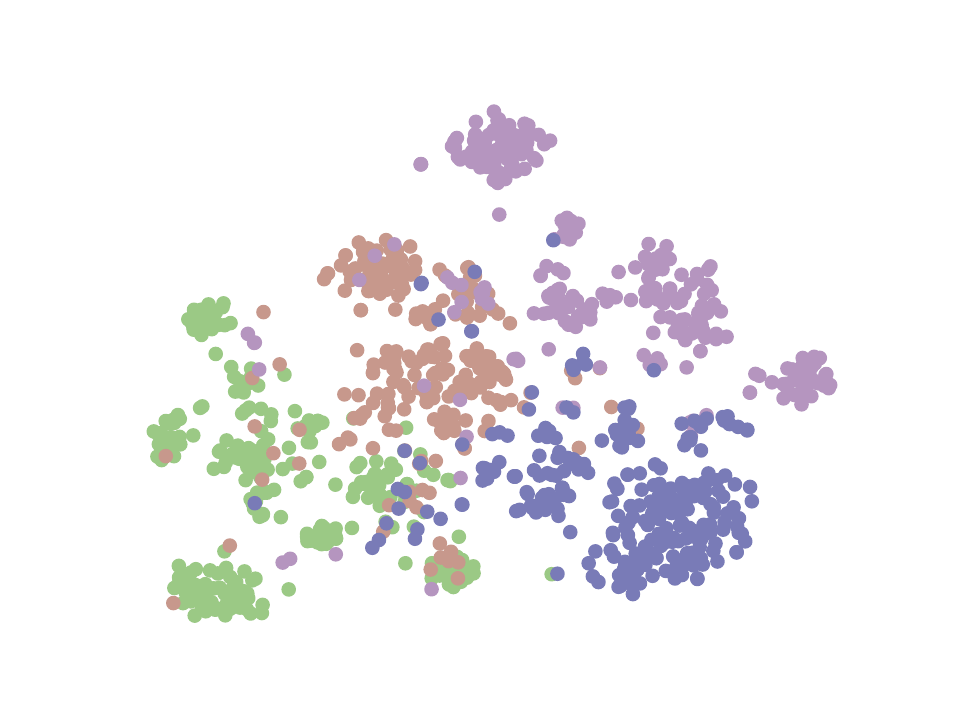}}
    \end{minipage}
    \begin{minipage}[b]{0.45\linewidth}
        \centering
        \subfloat[\small HeCo]{\includegraphics[width=\linewidth]{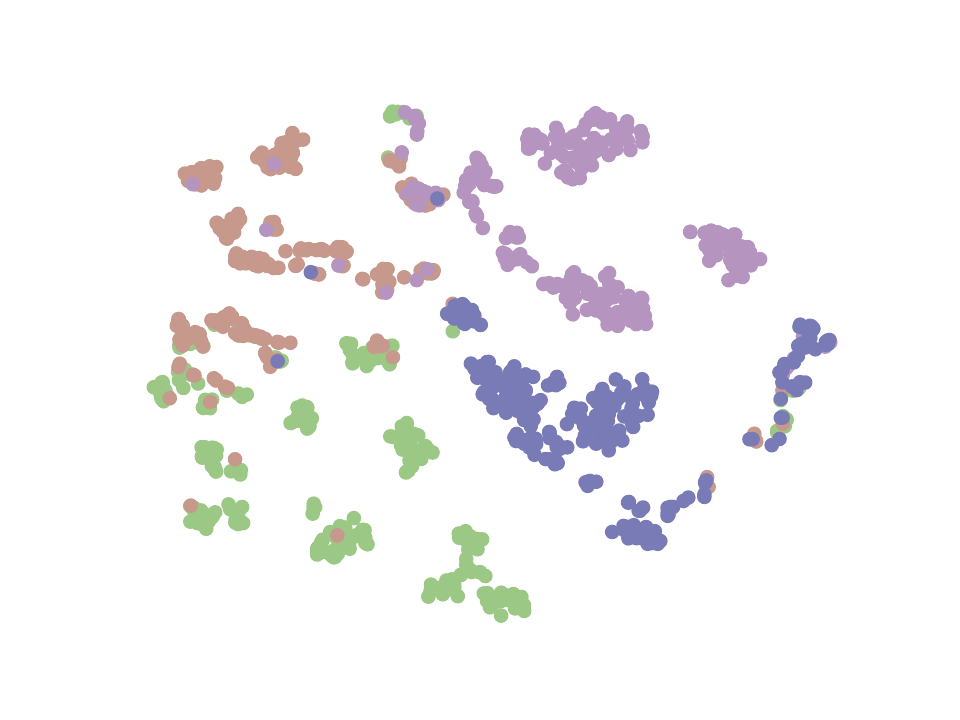}}
    \end{minipage}
    \begin{minipage}[b]{0.45\linewidth}
        \centering
        \subfloat[\small HGMAE]{\includegraphics[width=\linewidth]{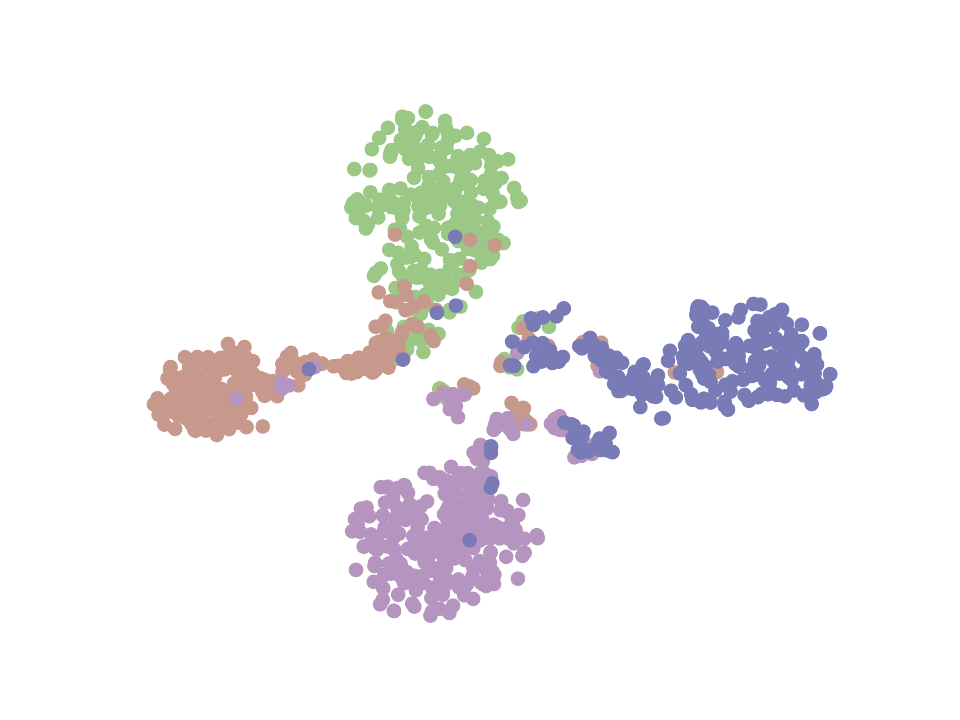}}
    \end{minipage}
    \begin{minipage}[b]{0.45\linewidth}
        \centering
        \subfloat[\small GC-HGNN]{\includegraphics[width=\linewidth]{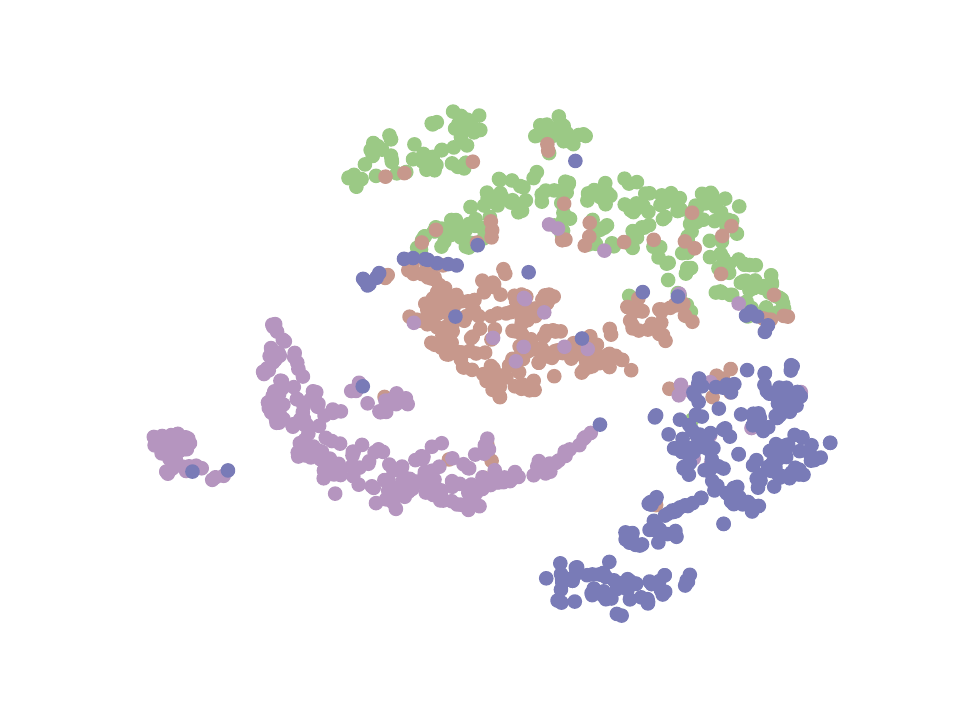}}
    \end{minipage}
    \caption{Embedding visualization of DBLP dataset. Different colors indicate different node category labels.}
    \label{fig:vis}
\end{figure}

\begin{figure*}[t]
    \centering
    \begin{minipage}[b]{0.23\linewidth}
        \centering
        \subfloat[\small ACM]{\includegraphics[width=\linewidth]{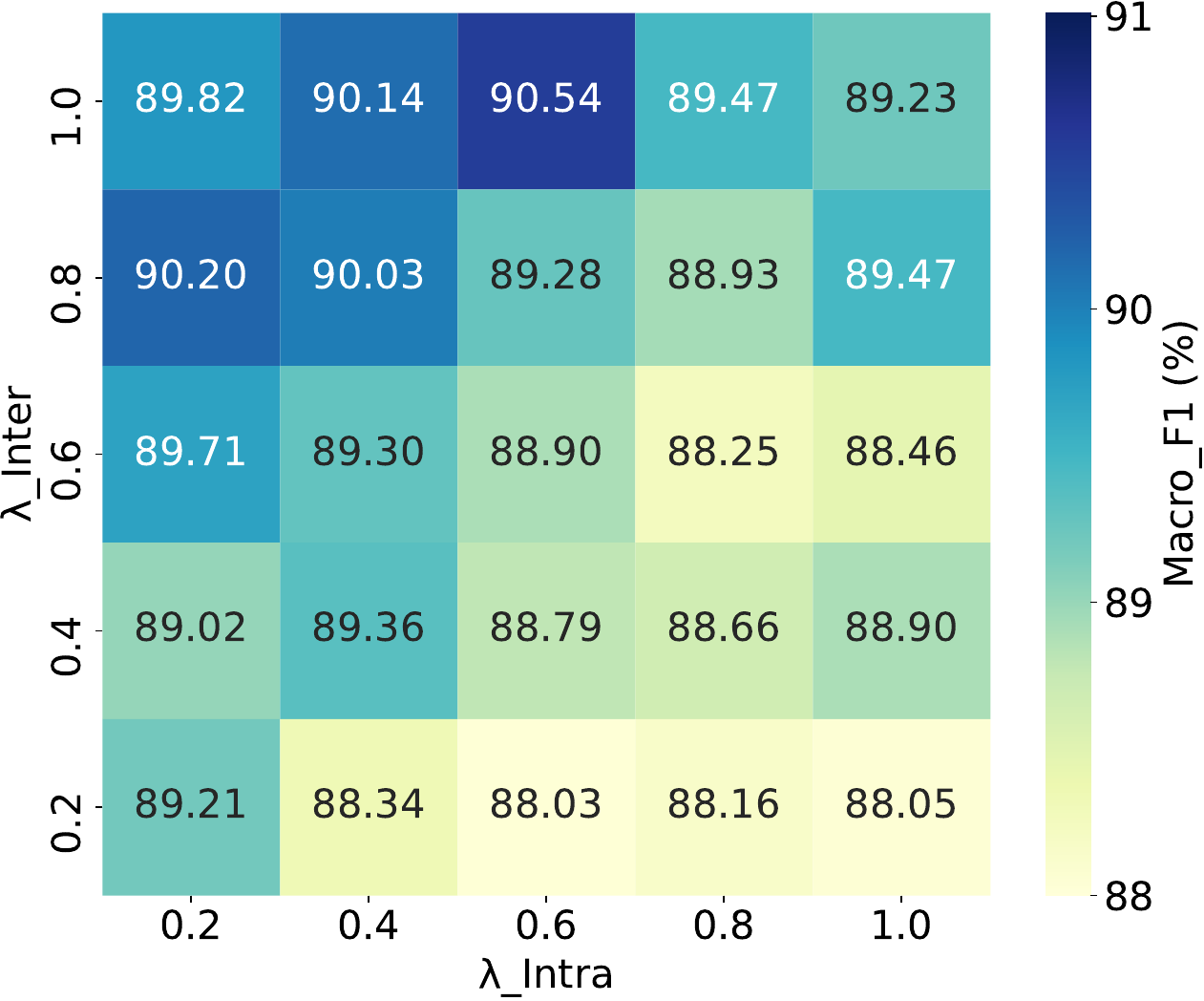}}
    \end{minipage}
    \begin{minipage}[b]{0.23\linewidth}
        \centering
        \subfloat[\small DBLP]{\includegraphics[width=\linewidth]{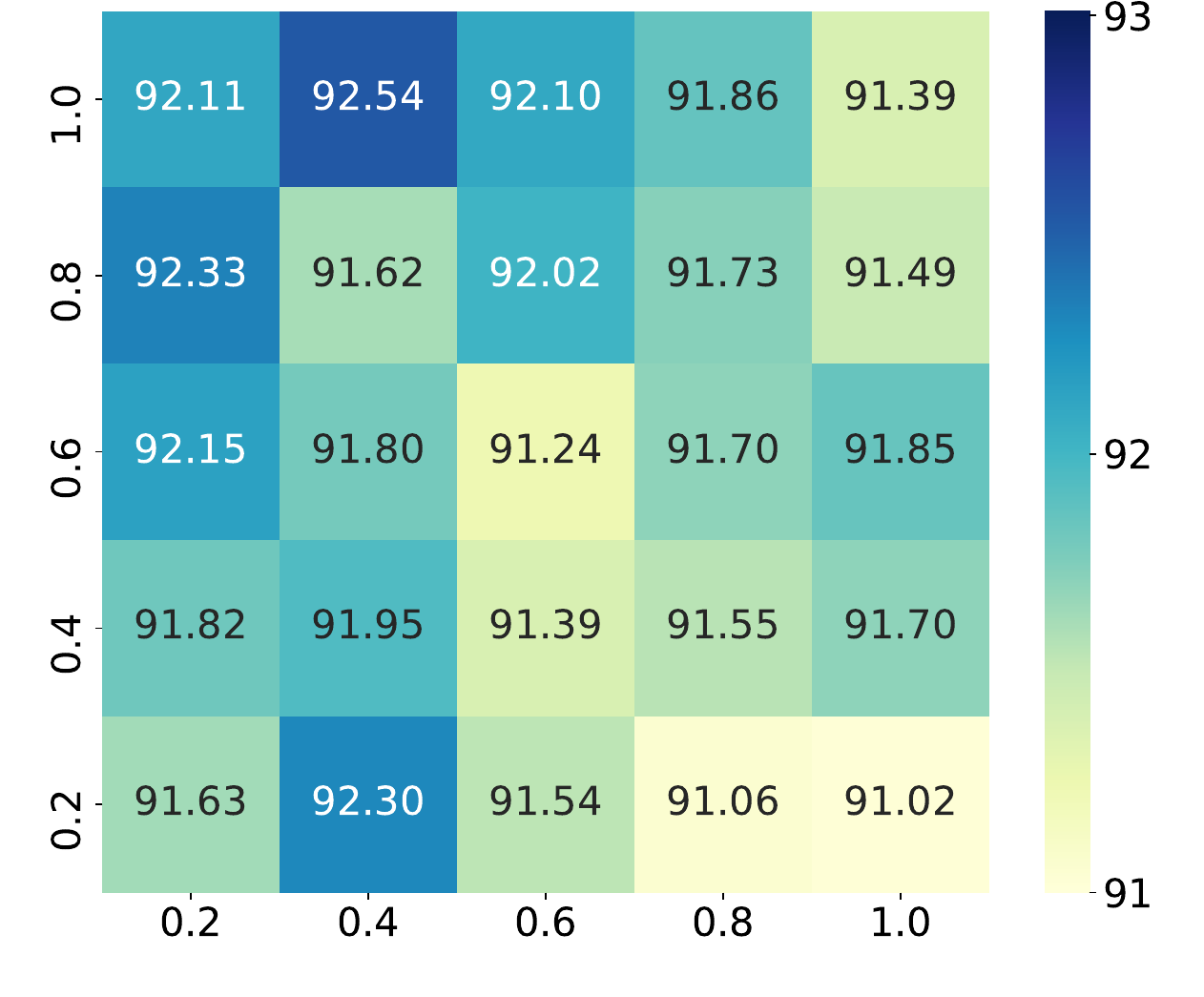}}
    \end{minipage}
    \begin{minipage}[b]{0.23\linewidth}
        \centering
        \subfloat[\small Aminer]{\includegraphics[width=\linewidth]{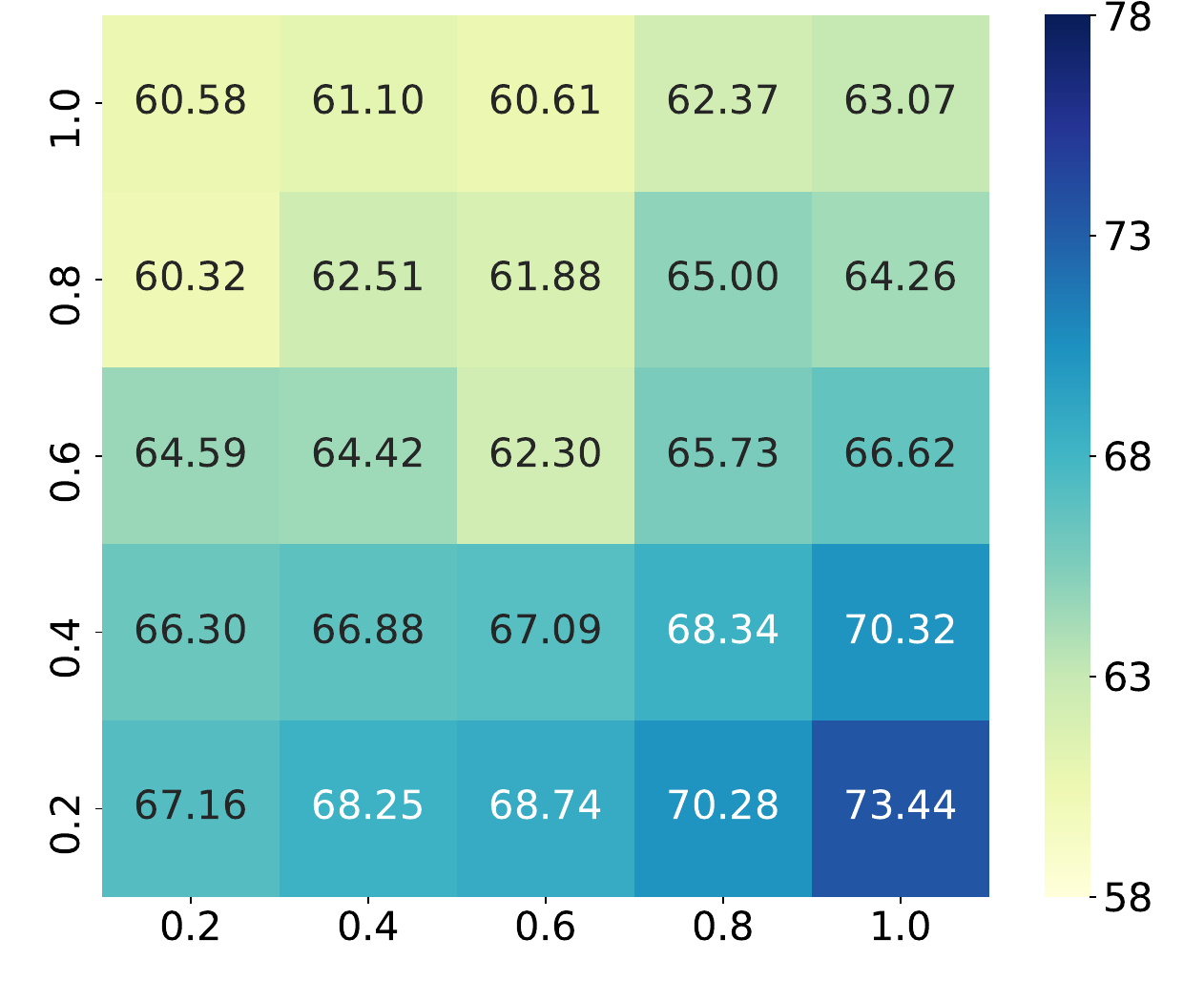}}
    \end{minipage}
    \begin{minipage}[b]{0.23\linewidth}
        \centering
        \subfloat[\small Freebase]{\includegraphics[width=\linewidth]{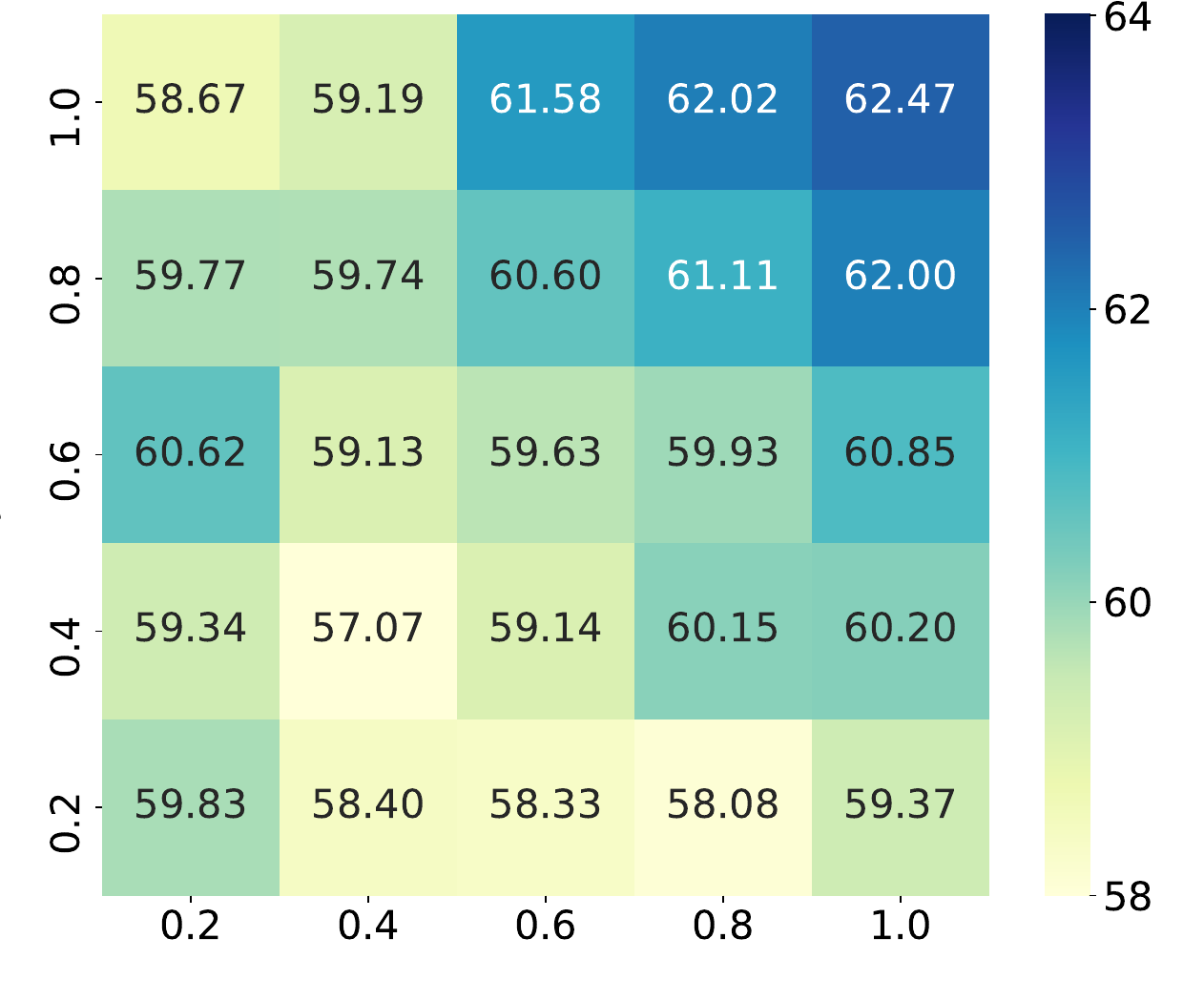}}
    \end{minipage}
    \caption{The impact wrt. hierarchical contrast coefficient. The axes on the left belong to the ACM and DBLP datasets, and the axes on the right belong to other datasets.}
    \label{fig7:hcl}
\end{figure*}

\subsection{Ablation Study(RQ2)}
In this section, we conducted ablation experiments to validate the effectiveness of key components in our GC-HGNN and provided possible explanations for the experimental results. We present three variants of our model.

\begin{itemize}[leftmargin=*]
\item$\boldsymbol{w} / \boldsymbol{o} {-} \mathbf{MAE}$:
This variant replaces the meta-path view's masked autoencoder (MAE) with a normal GAT \cite{2017gat} encoder. Because usually GAT has stronger performance than GCN \cite{2020lightgcn} on node classification tasks.
\item$\boldsymbol{w} / \boldsymbol{o} {-} \mathbf{SAM}$:
This variant replaces the sampling strategy (SAM) of GC-HGNN with random sampling while maintaining the same sampling ratio.
\item\textcolor{black}{$\boldsymbol{w} / \boldsymbol{o} {-} \mathbf{INTRA}$}:
\textcolor{black}{This variant replaces the intra-contrast corresponds to Eq. \ref{eq10}.}
\item\textcolor{black}{$\boldsymbol{w} / \boldsymbol{o} {-} \mathbf{INTER}$}:
\textcolor{black}{This variant cancels the inter-contrast integrated from Eqs. \ref{eq11} and \ref{eq12}.}
\item$\boldsymbol{w} / \boldsymbol{o} {-} \mathbf{HCL}$:
The last cancels the hierarchical contrastive learning (HCL) module and directly uses the InfoNCE \cite{2020simclr} loss for downstream tasks.
\end{itemize}

As shown in Figure \ref{fig4:ablation}, in all cases, removing the HCL module results in poorest performance compared to other variants. Additionally, variants where SAM or HCL is removed generally exhibit the worse performance in most cases. This suggests that inter-contrast through aligned local and global heterogeneous information significantly enhances semantics, while intra-contrast regulates semantic uniformity. We further highlights two important points: 1) The graph structural information is overemphasized, and reconstructing the graph links is not necessary; 2) The existing heterogeneous contrastive learning strategies are insufficient to serve as effective contrastive discriminators to enhance the masked autoencoder. \textcolor{black}{It is noteworthy that `INTER' serves as the principal loss function. This suggests that aligning high-order and first-order semantic representations of heterogeneous information can significantly improve the prediction accuracy, while contrast between meta-paths regulates semantic consistency \cite{sang2024intent}.}

\subsection{Interpretability and Visualization (RQ3)}
In this study, we examine the various modules of GC-HGNN, attempting to provide a rational explanation. We investigated the advantages of the generative-contrastive model and then further illustrated the motivation through the representation visualization.

\subsubsection{Interpretability}
Generative-contrastive (GC) representation learning \cite{2023gacn,2023vgcl,2020gan} employs discriminative loss functions as its objectives, aiming to address some inherent shortcomings of generative reconstruction, such as sensitive and conservative distribution issues and low-level abstraction objectives. In contrastive learning \cite{2020simclr, 2021heco, 2020moco, 2021sgl}, we can alleviate these issues by setting up appropriate contrastive discriminators to learn distinguishable information for different samples. However, the encoder and decoder in generative learning \cite{2018bert, 2022graphmae, 2023hgmae, 2023palm} provide powerful embedding expression capabilities. The decoder requires the representation to be reconstructable, meaning that it contains all the information of the input. Furthermore, GC-based self-supervised methods eschew the point-to-point reconstruction objective and aim to align distributions in the data. In summary, These methods \cite{2021ssl} integrate the advantages of both generative and contrastive methods, yet confront the challenge of constructing a framework within the field of heterogeneous graphs. We propose a masked autoencoder to enhance the expressive power of meta-path perspectives, followed by the use of  hierarchical contrastive learning and sampling strategies to serve as enhanced contrastive discriminators. Our experiments demonstrate that this is an effective and novel exploration.

\begin{figure*}[t]
    \begin{minipage}[b]{0.23\linewidth}
        \centering
        \subfloat[\small Macro-F1 wrt. dimension]{\includegraphics[width=\linewidth]{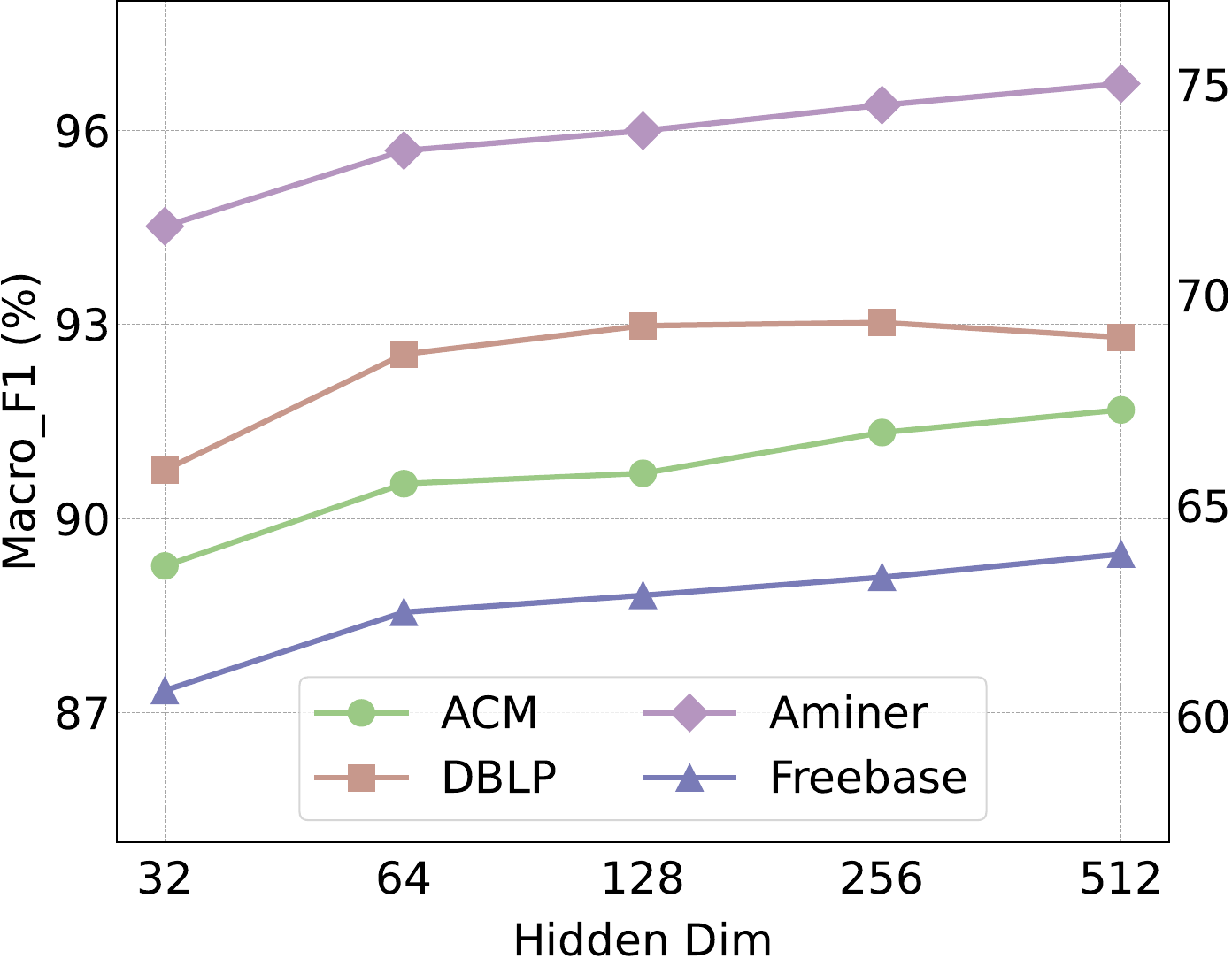}}
    \end{minipage}
    \begin{minipage}[b]{0.23\linewidth}
        \centering
        \subfloat[\small Micro-F1 wrt. dimension]{\includegraphics[width=\linewidth]{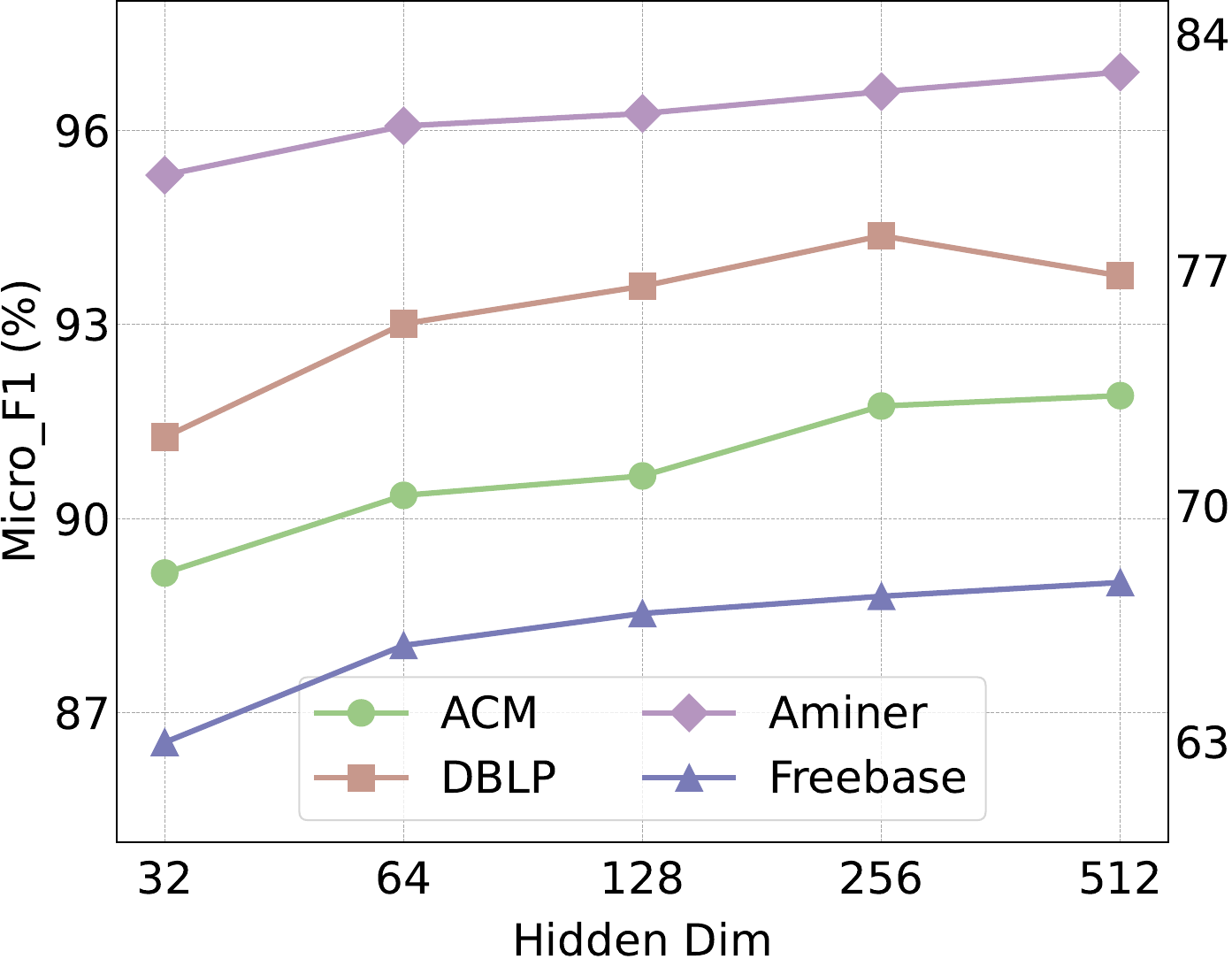}}
    \end{minipage}
    \begin{minipage}[b]{0.23\linewidth}
        \centering
        \subfloat[\small Macro-F1 wrt. mask ratio]{\includegraphics[width=\linewidth]{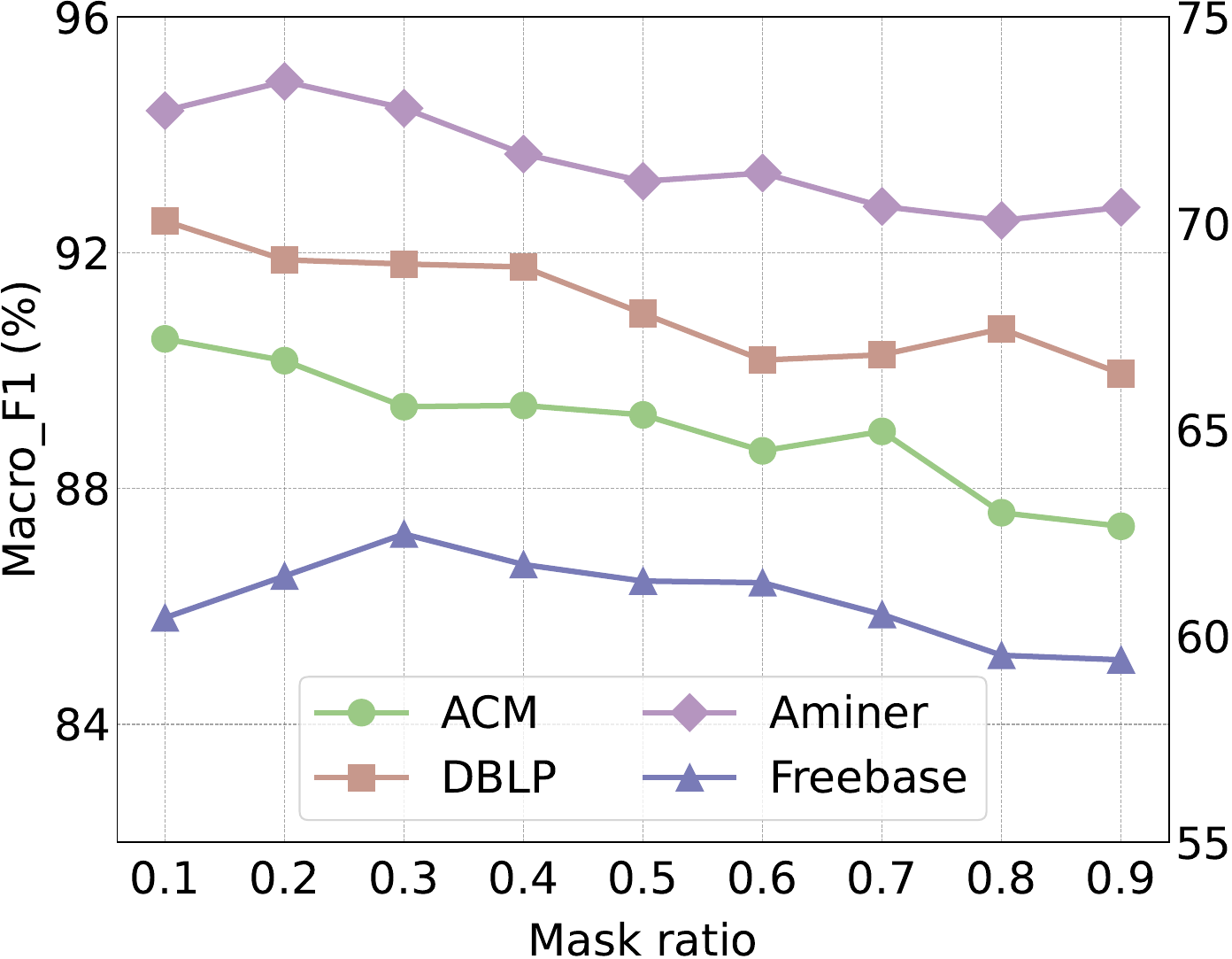}}
    \end{minipage}
    \begin{minipage}[b]{0.23\linewidth}
        \centering
        \subfloat[\small Micro-F1 wrt. mask ratio]{\includegraphics[width=\linewidth]{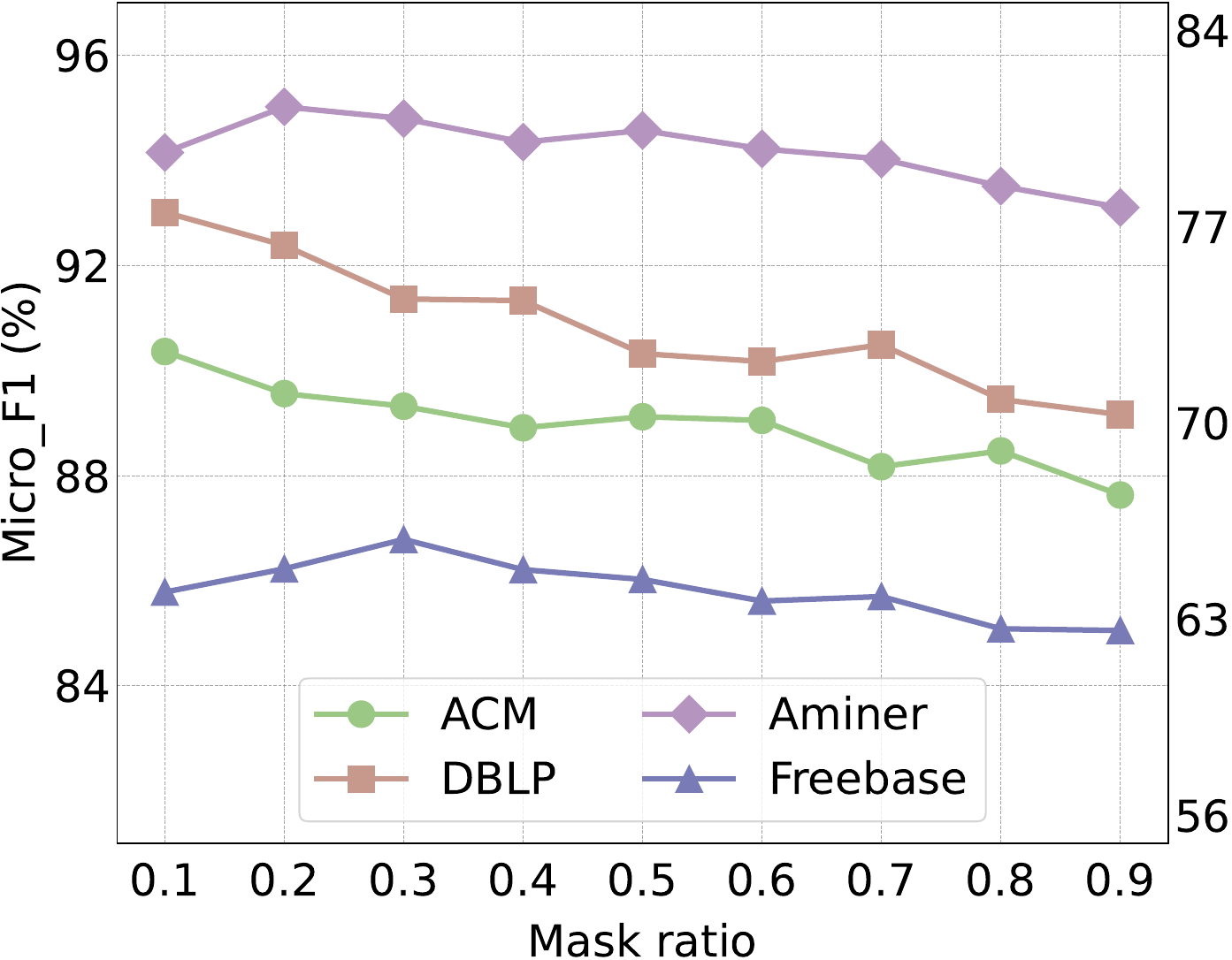}}
    \end{minipage}

    \caption{The impact wrt. dimension and mask ratio. The axes on the left belong to the ACM and DBLP datasets, and the axes on the right belong to other datasets.}
    \label{fig8:emb&mask}
\end{figure*}

\begin{table*}[t]
    \centering
    \captionsetup{justification=centering}
    \caption{Performance comparison with coefficient $\mathcal{L}_{Gen}$.}
    \begin{adjustbox}{width=0.95\textwidth}
    \begin{NiceTabular}{ccccccccc}
        \toprule
        \textbf{Dataset} & \multicolumn{2}{c}{\textbf{ACM}} & \multicolumn{2}{c}{\textbf{DBLP}} & \multicolumn{2}{c}{\textbf{Aminer}} & \multicolumn{2}{c}{\textbf{Freebase}} \\
        \cmidrule(lr){2-3} \cmidrule(lr){4-5} \cmidrule(lr){6-7} \cmidrule(lr){8-9}
        \textbf{Lambda} & Ma-F1 & Mi-F1 & Ma-F1 & Mi-F1 & Ma-F1 & Mi-F1 & Ma-F1 & Mi-F1 \\
        \midrule
        0.1 & \textbf{90.54$\pm$0.28} & \textbf{90.36$\pm$0.31} & \textbf{92.54$\pm$0.36} & \textbf{93.01$\pm$0.30} & 72.44$\pm$0.30 & 80.28$\pm$0.34 & 61.47$\pm$0.37 & 64.46$\pm$0.43 \\
        0.2 & 89.50$\pm$0.32 & 89.48$\pm$0.29 & 91.37$\pm$0.41 & 92.24$\pm$0.34 & \textbf{73.25$\pm$0.20} & \textbf{81.45$\pm$0.32} & \textbf{62.40$\pm$0.42} & \textbf{65.85$\pm$0.35} \\
        0.3 & 88.47$\pm$0.33 & 87.72$\pm$0.29 & 90.23$\pm$0.40 & 90.92$\pm$0.34 & 71.34$\pm$0.18 & 79.16$\pm$0.26 & 60.62$\pm$0.47 & 63.74$\pm$0.41 \\
        0.4 & 86.66$\pm$0.29 & 86.20$\pm$0.32 & 87.33$\pm$0.37 & 87.29$\pm$0.31 & 66.21$\pm$0.21 & 73.35$\pm$0.31 & 57.47$\pm$0.43 & 60.64$\pm$0.36 \\
        0.5 & 80.51$\pm$0.32 & 80.25$\pm$0.38 & 82.43$\pm$0.29 & 82.98$\pm$0.33 & 63.41$\pm$0.25 & 71.18$\pm$0.27 & 55.38$\pm$0.30 & 59.14$\pm$0.47 \\
        \bottomrule
    \end{NiceTabular}
    \end{adjustbox}
    \label{table4:loss_gem}
\end{table*}

\begin{table*}[h!]
    \centering
    \captionsetup{justification=centering}
    \caption{\textcolor{black}{Performance comparison of different encoder and decoder variants.}}
    \begin{adjustbox}{width=\textwidth}
    \color{black}
    \begin{NiceTabular}{c|c|cccccccc}
    \toprule[1pt]
    \multirow{2}{*}{ Encoder } & \multirow{2}{*}{ Decoder } & \multicolumn{2}{c}{ ACM } & \multicolumn{2}{c}{ DBLP } & \multicolumn{2}{c}{ Aminer } & \multicolumn{2}{c}{ Freebase } \\
    \cmidrule{3-10} & & Macro-F1 & Micro-F1 & Macro-F1 & Micro-F1 & Macro-F1 & Micro-F1 & Macro-F1 & Micro-F1 \\
    \midrule 
    GAT & GAT & $\underline{90.29 \pm 0.22}$ & $\underline{90.16 \pm 0.30}$ & $\underline{92.14 \pm 0.32}$ & $\underline{92.49 \pm 0.41}$ & $\mathbf{73.44} \pm \mathbf{0.20}$ & $\mathbf{81.28} \pm \mathbf{0.32}$ & $\mathbf{62.47} \pm \mathbf{0.42}$ & $\mathbf{65.85} \pm \mathbf{0.35}$ \\
    GCN & GCN & $\mathbf{90.54} \pm \mathbf{0.28}$ & $\mathbf{90.36} \pm \mathbf{0.31}$ & $91.16 \pm 0.24$ & $91.52 \pm 0.27$ & $71.90 \pm 0.32$ & $79.48 \pm 0.43$ & $61.85 \pm 0.20$ & $64.88 \pm 0.24$ \\
    GAT & GCN & $89.24 \pm 0.27$ & $89.46 \pm 0.32$ & $90.74 \pm 0.18$ & $91.15 \pm 0.23$ & $\underline{72.69 \pm 0.25}$ & $\underline{80.44 \pm 0.34}$ & $61.23 \pm 0.33$ & $63.78 \pm 0.26$ \\
    GCN & GAT & $89.68 \pm 0.18$ & $89.61 \pm 0.36$ & $\mathbf{92.54} \pm \mathbf{0.36}$ & $\mathbf{93.01} \pm \mathbf{0.30}$ & $72.31 \pm 0.44$ & $79.98 \pm 0.21$ & $\underline{62.02 \pm 0.29}$ & $\underline{65.24 \pm 0.35}$ \\
    \bottomrule[1pt]
    \end{NiceTabular}
    \end{adjustbox}
    \label{table:1.1}
\end{table*}

\subsubsection{Visualization} 
To gain a more intuitive understanding of the embeddings from different models, we employ t-SNE \cite{2008tsne} to reduce the dimensionality of the embeddings and visualize them with labels. As illustrated in Figure \ref{fig:vis}, we select three representative models: unsupervised MP2vec \cite{2017metapath2vec}, contrastive HeCo \cite{2021heco}, and generative HGMAE \cite{2023hgmae}. We can observe that MP2vec performs the worst, with different types of nodes mixed together, making them difficult to distinguish. HeCo effectively differentiate categories, but each block remains separated and does not form distinct clusters. HGMAE uses a hidden dimension of 256 and is a generative method, thus exhibiting more evident clustering phenomena. However, our GC-HGNN, based on a contrastive framework, demonstrates both clustering and uniform distribution, effectively balancing the advantages of contrastive and generative approaches. This phenomenon further indicates the effectiveness of our GC-based paradigm.

\subsection{Parameter Sensitivity (RQ4)}
In this study, we explore the impact of key hyperparameters on heterogeneous graph node classification. Moreover, we investigate the effects of hierarchical contrast coefficient, embedding dimension, and mask ratio. 

\begin{itemize}[leftmargin=*]
    \item \textbf{Hierarchical contrast coefficient.}
    The $\mathcal{L}_{hcl}$ contains $\lambda_{{Intra}}$ and $\lambda_{{Inter}}$, which are set from 0 to 1 with an incremental step. In Figure \ref{fig7:hcl}, it is observed that the performance improves as $\lambda_{{Inter}}$ increases. This phenomenon aligns with the principle of our proposed sampling method, where inter-contrast has more hard negative samples.
    \item \textbf{Generative coefficient.}
    In Table \ref{table4:loss_gem}, we evaluated $\mathcal{L}_{Gen}$ ranging from 0.1 to 0.5, observing that performance metrics typically decline as the $\mathcal{L}_{Gen}$ increases. It can be observed that the model tends to favor a smaller (generative learning) GL loss, which aligns with our idea regarding the auxiliary role of GL in the (generative-contrastive) GC paradigm.
    \item \textbf{Embedding dimension.}
    We examine how different embedding sizes affect node classification performance ranging from 32 to 512. Figure \ref{fig8:emb&mask} shows that most datasets show improvement up to a 512-dimensional embedding, apart from DBLP, which peaks at 256. 
    However, to balance the marginal gains against higher computational costs and overfitting risks, we standardize the embedding size at 64 for all tests.
    \item \textbf{Mask ratio.} 
    We utilize a masked autoencoder, where the mask ratio plays a crucial role as a hyper-parameter in generating negative samples. As shown in Figure \ref{fig8:emb&mask}, our model achieves better results with a lower mask ratio, contrary to the conclusions drawn by HGMAE \cite{2023hgmae}. 
    We attribute this to the fact that, compared to generative feature reconstruction, the game between contrastive learning and generation learns to reconstruct the original data distribution by minimizing the distributional divergence rather than reconstructing individual samples \cite{2021ssl}.
    \item \textcolor{black}{\textbf{Encoder types in MAE.}
    We set multiple variants for the encoder and decoder to test their effect on GC-HGNN. Theoretically, our encoder can be any GNN model (such as GIN \cite{2018gin}), or even an MLP. As shown in Table \ref{table:1.1}, the best result is highlighted in bold, while the second-best is underlined. We choose a combination of the commonly used GAT \cite{2017gat} and GCN \cite{2016gcn}. It can be observed that the GAT-GAT combination achieves the best or second-best performance on most datasets. This phenomenon aligns with the conclusion that the attention mechanism can effectively capture relationships between nodes in node classification tasks. Moreover, all variants outperform the MAE-free variant from the ablation study, demonstrating that MAE effectively mitigates noise limitation.}
\end{itemize}
In general, the findings suggest that higher inter-contrast, higher embedding dimensions, and lower mask ratios typically yield better results, with some dataset-specific variations.

\section{\textcolor{black}{Conclusion and Future Work}}
In this paper, we proposed a novel generative-contrastive heterogeneous graph neural network termed GC-HGNN. It represented a generative learning enhanced contrastive paradigm and a meaningful exploration. We combined meta-paths and masked autoencoders to create augmented contrastive views without disrupting the data structure. The generative-contrastive framework significantly improves the performance of node classification and link prediction tasks. We compared our model with seventeen state-of-the-art models on eight real-world datasets, demonstrating that it outperformed the them on these tasks.


\section*{Acknowledgment}
This work is supported by the National Science Foundation of China (No. 62206002, No. 62272001 and No. 62206004), Hefei Key Common Technology Project (NO. 2023SGJ014), Natural Science Foundation of Anhui Province (No. 2208085QF195) and Xunfei Zhiyuan Digital Transformation Innovation Research Special for Universities (NO. 2023ZY001).
\nocite{*}
\bibliographystyle{IEEEtran}
\bibliography{ref}
\end{document}